\documentclass[lettersize,journal]{IEEEtran}
\usepackage{amsmath,amsfonts}
\usepackage{algorithmic}
\usepackage{algorithm}
\usepackage{array}
\usepackage[caption=false,font=normalsize,labelfont=sf,textfont=sf]{subfig}
\usepackage{textcomp}
\usepackage{stfloats}
\usepackage{url}
\usepackage{verbatim}
\usepackage{graphicx}
\usepackage{cite}
\usepackage{xspace}
\usepackage{xfrac}
\usepackage{booktabs}
\usepackage{makecell}
\usepackage{amssymb}
\usepackage[normalem]{ulem}
\useunder{\uline}{\ul}{}
\makeatletter
\DeclareRobustCommand\onedot{\futurelet\@let@token\@onedot}
\def\@onedot{\ifx\@let@token.\else.\null\fi\xspace}

\def\ie{\emph{i.e}\onedot} 
 
\def\etc{\emph{etc}\onedot} 
 
\def\etal{\emph{et al}\onedot}
\makeatother

\begin{document}

\title{Learning Inverse Laplacian Pyramid\\for Progressive Depth Completion}

\author{Kun Wang, Zhiqiang Yan, Junkai Fan, Jun Li and Jian Yang
\thanks{All authors are with PCA Lab, Key Lab of Intelligent Perception and Systems for High-Dimensional Information of Ministry of Education, and Jiangsu Key Lab of Image and Video Understanding for Social Security, School of Computer Science and Engineering, Nanjing University of Science and Technology, Nanjing, China. Email: \{kunwang, yanzq, junkai.fan, junli, csjyang\}@njust.edu.cn. Jun Li and Jian Yang are corresponding authors.}
}

\markboth{Journal of \LaTeX\ Class Files,~Vol.~14, No.~8, August~2021}%
{Shell \MakeLowercase{\textit{et al.}}: A Sample Article Using IEEEtran.cls for IEEE Journals}

\IEEEpubid{0000--0000/00\$00.00~\copyright~2021 IEEE}

\maketitle

\begin{abstract}

Depth completion endeavors to reconstruct a dense depth map from sparse depth measurements, leveraging the information provided by a corresponding color image. Existing approaches mostly hinge on single-scale propagation strategies that iteratively ameliorate initial coarse depth estimates through pixel-level message passing. Despite their commendable outcomes, these techniques are frequently hampered by computational inefficiencies and a limited grasp of scene context. To circumvent these challenges, we introduce LP-Net, an innovative framework that implements a multi-scale, progressive prediction paradigm based on Laplacian Pyramid decomposition. Diverging from propagation-based approaches, LP-Net initiates with a rudimentary, low-resolution depth prediction to encapsulate the global scene context, subsequently refining this through successive upsampling and the reinstatement of high-frequency details at incremental scales. We have developed two novel modules to bolster this strategy: 1) the Multi-path Feature Pyramid module, which segregates feature maps into discrete pathways, employing multi-scale transformations to amalgamate comprehensive spatial information, and 2) the Selective Depth Filtering module, which dynamically learns to apply both smoothness and sharpness filters to judiciously mitigate noise while accentuating intricate details. By integrating these advancements, LP-Net not only secures state-of-the-art (SOTA) performance across both outdoor and indoor benchmarks such as KITTI, NYUv2, and TOFDC, but also demonstrates superior computational efficiency. At the time of submission, LP-Net ranks 1st among all peer-reviewed methods on the official KITTI leaderboard. The source code will be made publicly accessible upon paper acceptance.

\end{abstract}


\section{Introduction}

\IEEEPARstart{M}{onocular} depth perception entails estimating the distance from object surfaces to the camera plane. This task is pivotal across numerous domains, attracting substantial interest from both industry and academia, as comprehensive and precise depth information can markedly enhance applications such as 3D reconstruction \cite{recon_tcsvt, altnerf}, augmented reality \cite{ar_tcsvt, ar2017}, and robotics \cite{robot_tcsvt, dong2022robot}, among others. A segment of research has concentrated on directly predicting depth from a single image, known as Monocular Depth Estimation (MDE) \cite{mde_tcsvt, dcdepth, rnw, fan2025depth, wang2024sgnet}. However, MDE is fundamentally ill-posed, plagued by scale ambiguity, which curtails its accuracy. Conversely, existing range sensors like LiDAR and Time-of-Flight (ToF) devices encounter inherent constraints, including sparse measurements and reduced robustness, impeding their capacity to yield dense, high-quality depth data. In response to these limitations, depth completion has emerged, aiming to amalgamate the merits of both image-based and sparse measurement modalities to efficaciously generate dense, precise depth maps for various downstream tasks and applications.

\begin{figure}
    \centering
    \includegraphics[width=0.95\linewidth]{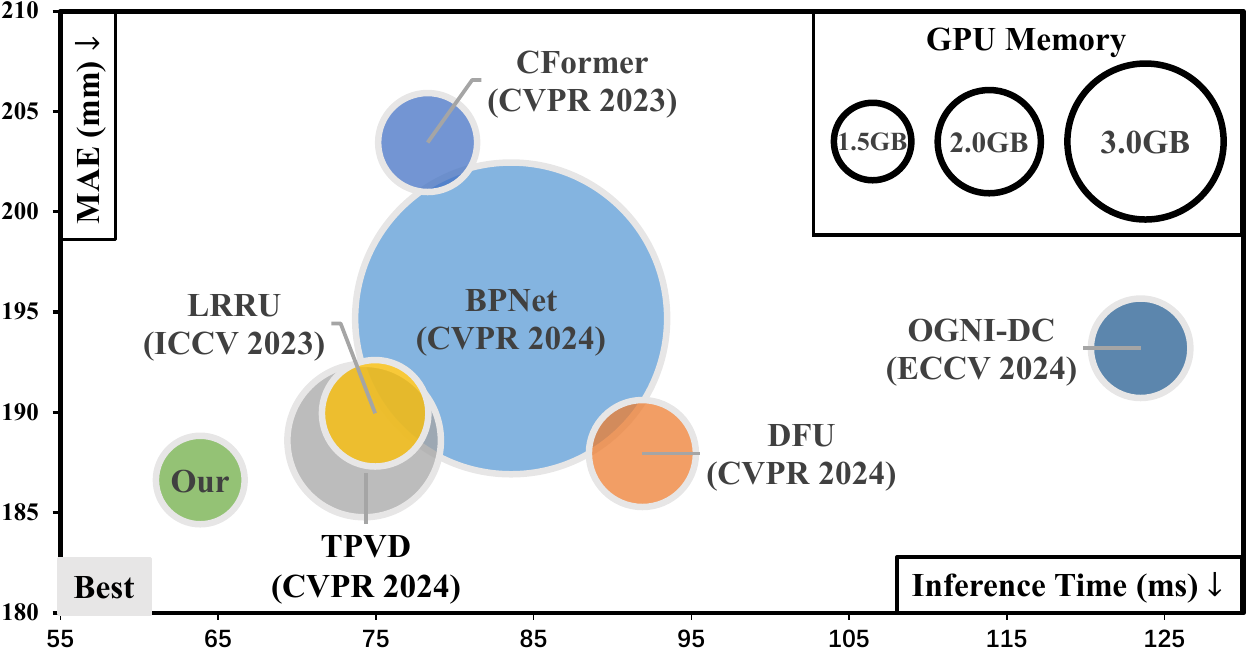}
    \caption{Visual comparison with recent state-of-the-art (SOTA) methods on prediction accuracy and computational efficiency. MAE metrics are sourced from the KITTI online leaderboard, while inference time and GPU memory usage are assessed on $1216\times 256$ images using a single RTX 4090 GPU. Our proposed LP-Net demonstrates superior performance over previous SOTA methods in both accuracy and efficiency.}
    \label{fig.1}
\end{figure} 

\begin{figure}
    \centering
    \includegraphics[width=0.98\linewidth]{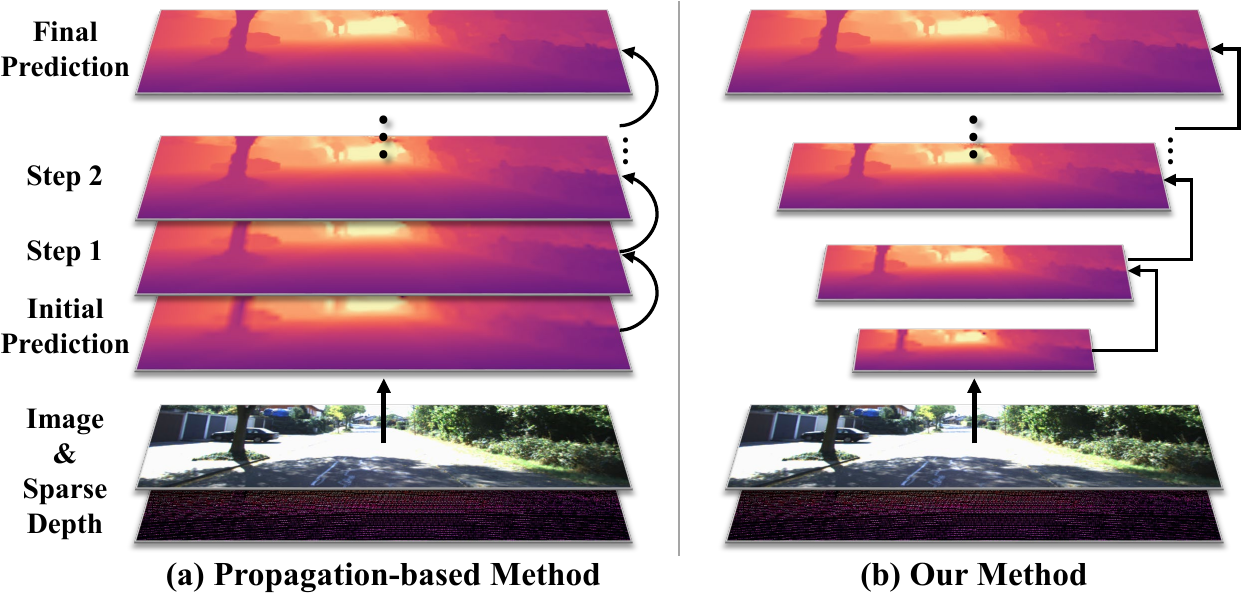}
    \caption{Visual comparison between existing single-scale, propagation-based approaches and our multi-scale, Laplacian Pyramid-based prediction scheme. (a) Propagation-based methods iteratively refine an initial coarse depth prediction through recurrent convolution, aggregating depth information from neighboring pixels. (b) Our proposed prediction scheme progressively upsamples the initial low-resolution prediction and recovers high-frequency details at each scale with a novel selective filtering mechanism, facilitating both accurate and efficient depth completion.}
    \label{fig.prop}
\end{figure}

\IEEEpubidadjcol
Early depth completion approaches rooted in conventional techniques such as superpixels \cite{van2013depth}, image filtering \cite{ku2018defense}, and optimization strategies \cite{zhang2018deep}. Despite some successes, these methods were constrained by both accuracy and scalability. The advent of deep learning has significantly propelled this field, fostering the development of more robust, data-driven solutions. Among these, single-scale propagation-based methods \cite{cspn, nlspn, wang2023decomposed, rignet} have emerged as the prevailing approach in recent SOTA techniques. Here, propagation denotes a message-passing process employing a recurrent convolutional operation to aggregate depth information for each pixel from its neighbors, based on a predicted affinity matrix. Neighbors may be either fixed \cite{cspn++} or adaptively chosen \cite{lin2023dyspn}. Through iterative operations, an initially coarse depth prediction is progressively refined, enhancing depth accuracy, as illustrated in Fig. \ref{fig.prop} (a). However, these methods grapple with two fundamental drawbacks: 1) the spatial redundancy inherent in depth maps renders the single-scale, pixel-wise propagation computationally inefficient. 2) the limited contextual information provided by individual pixels necessitates numerous iterations for comprehensive global scene comprehension, often resulting in inadequate global context during propagation. As a result, the above shortcomings limit the overall performance and efficiency of existing methods, as illustrated in Fig. \ref{fig.1}.

In this paper, we introduce LP-Net, an innovative framework designed to overcome the previously discussed limitations. Diverging from conventional single-scale, propagation-based approaches, LP-Net implements a multi-scale, progressive prediction scheme based on Laplacian Pyramid decomposition. The Laplacian Pyramid serves as a hierarchical image representation technique comprising a series of bandpass images of progressively diminishing resolution alongside a low-frequency residual. Each bandpass image captures the differences between successive resolution levels of the original image, encapsulating the details lost in downsampling. The low-frequency residual is essentially the smallest scale version of the image. LP-Net inverts this process, starting with the estimation of the low-frequency residual to encapsulate global scene context. Subsequently, the initial depth map is progressively upsampled, and a novel selective filtering methodology is introduced to reinstate the high-frequency details lost at each scale. The process is visually illustrated in Fig. \ref{fig.prop} (b). This strategic approach from global to local reduces the complexity of depth completion while boosting prediction efficiency. To realize our method, we introduce two pivotal modules: the Multi-path Feature Pyramid (MFP) module and the Selective Depth Filtering (SDF) module. The MFP module bifurcates the feature map into multiple pathways, each subjected to unique transformations to produce representations at various visual scales. These are then upsampled to align with a uniform resolution and amalgamated with the original feature map, enhancing global context perception. The SDF module dynamically learns to deploy both a smoothness filter and a sharpness filter, selectively applied to the depth map to effectively integrate sparse depth measurements, suppress noises, and accentuate fine-gained details.

Our main contributions can be delineated as follows:
\begin{itemize}
    \item We introduce LP-Net, a novel depth completion framework that mitigates the limitations of existing single-scale propagation-based methods through a multi-scale, progressive prediction scheme based on the Laplacian Pyramid decomposition. This strategy facilitates both a comprehensive scene understanding and an efficient, hierarchical depth refinement process.
    \item We devise the MFP module, which partitions feature maps into multiple pathways, applying multi-scale transformations to bolster the model's global perception capabilities. Furthermore, we develop the SDF module, which dynamically learns both a smoothness filter and a sharpness filter to judiciously integrate sparse depth measurements, mitigate noises, and accentuate local details.
    \item By integrating these innovations, LP-Net achieves SOTA performance on both outdoor and indoor benchmarks including KITTI, NYUv2, and TOFDC, while also exhibiting superior computational efficiency. Notably, at the time of submission, LP-Net ranks 1st among all peer-reviewed methods on the online KITTI leaderboard.
\end{itemize}

\section{Related Work}

\paragraph{Traditional Methods}
The depth completion task involves reconstructing a dense, accurate depth map from sparse measurements. These sparse inputs can originate from range sensors like LiDAR, Time-of-Flight (ToF), and structured light, or from computational techniques such as Structure-from-Motion (SfM) and binocular stereo. Traditional methods for depth completion often employ conventional techniques to interpolate missing depth values. For instance, Depth SEEDS \cite{depth_seeds} leveraged an extended version of the SEEDS \cite{van2015seeds} superpixel technique, incorporating available depth measurements to recover absent depth information. IP-Basic \cite{ip_basic} applied a series of traditional image processing operations, including dilation, closure, bilateral filtering, and hole filling, \etc, to densify sparse depth maps. Zhang \etal \cite{zhang2018deep} utilized surface normal and occlusion boundary priors to solve for a dense depth map from sparse observations by minimizing an objective function that balances geometric and smoothness constraints. Zhao \etal \cite{zhao2021surface} introduced a surface geometry model enhanced by an outlier removal algorithm to fill the missing depth values based on a local surface sharing assumption. 

\paragraph{Deep Methods}
The development of deep learning has shifted the focus towards neural network-based, data-driven approaches to enhance prediction accuracy and scalability. Handling sparse and irregularly spaced input data presents a significant challenge in depth completion. To tackle this, Uhrig \etal \cite{uhrig2017sparsity} introduced a sparse invariant convolutional operator that explicitly accounts for missing values during convolution, thus better processing irregularly spaced sparse depths. Huang \etal \cite{huang2019hms} enhanced this with HMS-Net, an encoder-decoder architecture employing novel sparsity-invariant operations to fuse multi-scale features from various layers. Eldesokey \etal \cite{eldesokey2020uncertainty} introduced an algebraically-constrained normalized convolution layer for managing highly sparse depth data. Additionally, some approaches adopt a multi-stage framework, initially filling in sparse depth map values to circumvent direct convolution on sparse data. Chen \etal \cite{chen2018estimating} employed nearest interpolation on the sparse depth map before feeding it into a deep network; Liu \etal \cite{liu2021learning} used a differentiable kernel regression layer in place of manual interpolation, and BP-Net \cite{bp_net} aggregated information from nearest neighbors for filling missing depth values through bilateral propagation.

\paragraph{Multi-modal Fusion}
Another research avenue emphasizes the effective integration of features from both RGB images and sparse depth maps. Chen \etal \cite{chen2019learning} proposed to fuse 2D and 3D features extracted through 2D convolutions on images and continuous convolutions on 3D points, respectively. DeepLiDAR \cite{deeplidar} used surface normals as an intermediate representation for fusion with sparse depth. GuideNet \cite{guidenet} utilized features from both modalities as guidance to generate context-dependent, spatially-variant convolution kernels for adaptive feature extraction from sparse depth maps. RigNet \cite{rignet} and Wang \etal \cite{wang2023decomposed} further improved \cite{guidenet} by introducing repetitive architecture and novel decomposition schemes, respectively. Recent works have also explored attention mechanisms for multi-modal feature fusion and innovative projection views for enhancing geometric representation capabilities. For example, GuideFormer \cite{guideformer} introduced a guided-attention module for multi-modal fusion, BEV@DC \cite{bev_dc} employed a bird's-eye view for training, leveraging the rich geometric details of LiDAR, whereas TPVD \cite{tpvd} used a tri-perspective view decomposition to model 3D geometry explicitly.

\begin{figure*}
    \centering
    \includegraphics[width=0.98\linewidth]{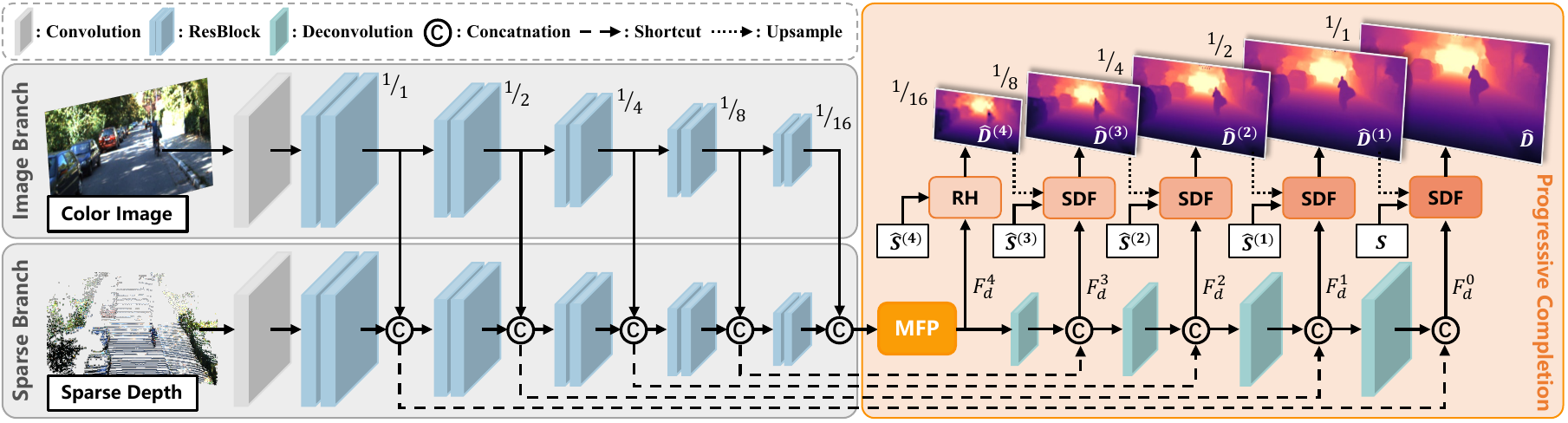}
    \caption{Overall Framework of LP-Net. MFP, RH and SDF stand for the Multi-path Feature Pyramid module, Regression Head and Selective Depth Filtering module, respectively. $S$ represents the input sparse depth, while $\hat{S}^{(1)}\sim \hat{S}^{(4)}$ are its progressively lower-resolution versions, obtained through a weighted pooling operation. $F_{d}^{0}\sim F_{d}^{4}$ indicate the decoder feature maps. The prediction of the final depth map $\hat{D}$ is structured into five progressive steps, beginning with a direct regression and confidence-based fusion with $\hat{S}^{(4)}$ to produce the low-frequency residual $\hat{D}^{(4)}$. Subsequently, $\hat{D}^{(4)}$ undergoes iterative upsampling, fusion with the corresponding sparse measurements, and refinement via the SDF module to yield more accurate, higher-resolution depth maps.}
    \label{fig.framework}
\end{figure*} 

\paragraph{Propagation Methods}
Propagation-based methods have significantly advanced recent depth completion strategies. CSPN \cite{cspn} introduced a convolutional spatial propagation network, initiating with a coarse depth prediction and progressively refining it through learned affinity among fixed neighbors in a recurrent convolutional process. CSPN++ \cite{cspn++} extent this by combining results from various kernel sizes and iteration steps. NLSPN \cite{nlspn} improved upon this by learning to select relevant neighbors through predicted offsets from a regular grid. DySPN \cite{dyspn, lin2023dyspn} shifted to an attention-based dynamic approach to learn affinity, addressing issues like fixed affinity and over-smoothing. GraphCSPN \cite{graphcspn} integrated 3D information into a graph neural network to estimate neighbors during updates, and LRRU \cite{lrru} introduced a lightweight network that adjusts kernel scope from large to small to capture dependencies from long to short range. Despite their wide adoption for post-processing to enhance prediction accuracy, these propagation-based methods are often computationally intensive and lack a comprehensive understanding of global scenes. In contrast, our proposed method constructs a multi-scale scheme that progressively upsamples initial predictions and recovers high-frequency details at each scale using a novel selective filtering technique, thereby achieving state-of-the-art performance and high computational efficiency.

\section{Method}

In this section, we present our LP-Net framework, which utilizes a multi-scale, progressive depth completion scheme founded on Laplacian Pyramid. We commence with a succinct overview of Laplacian Pyramid decomposition as a foundational concept. Following this, we explore the progressive prediction scheme in detail and elucidate the architecture of the network along with its constituent modules. Lastly, we describe the loss function utilized for training our model.

\subsection{Reviewing Laplacian Pyramid}

The Laplacian Pyramid is a bandpass pyramid technique used for multi-scale image representation, specifically tailored to capture the high-frequency or detail components of images across different scales. It consists of a series of bandpass images with progressively lower resolution, alongside a low-frequency residual. At each scale, these bandpass images capture the differences between successive resolution levels of the original image, thus encapsulating the details lost in downsampling. The low-frequency residual corresponds to the smallest scale version of the image, embodying its global, low-frequency structure. We denote the operations of upsampling and downsampling by $\text{up}(\cdot)$ and $\text{down}(\cdot)$, respectively. As an example, a three-level Laplacian Pyramid decomposition of an image $x$ can be formally described as follows:
\begin{equation}
    \begin{aligned}
        x^{(3)}&=\text{down}(\text{down}(x)),\\
        x^{(2)}&=\text{down}(x)-\text{up}(x^{(3)}),\\
        x^{(1)}&=x-\text{up}(\text{down}(x)),
    \end{aligned}
\end{equation}
where $x^{(3)}$ is the lowest-resolution image, representing the low-frequency residual, while $x^{(2)}$ and $x^{(1)}$ are bandpass images at full and half resolution, respectively. The original image $x$ can be effectively reconstructed by reversing this process, which is expressed as:
\begin{equation}
    x = \text{up}(\text{up}(x^{(3)}) + x^{(2)}) + x^{(1)}.
    \label{eq.lp_recon}
\end{equation}
Our proposed LP-Net mirrors the inverse process of the Laplacian Pyramid to reconstruct dense depth maps, progressing from a global perspective to local details. This hierarchical prediction scheme reduces the complexity of depth completion while enhancing computational efficiency.

\subsection{Depth Completion Based on Laplacian Pyramid}

\begin{figure*}
    \centering
    \includegraphics[width=0.98\linewidth]{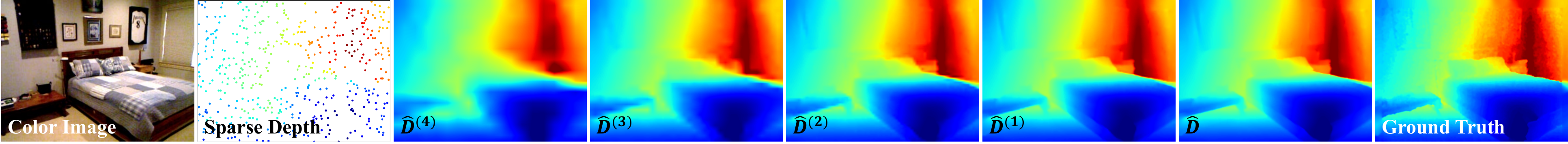}
    \caption{Evolution of depth completion results. We illustrate the progression of our depth completion scheme by showcasing the intermediate results on the NYUv2 dataset. The depth predictions $\hat{D}^{(4)}\sim \hat{D}^{(1)}$ have been upsampled to full resolution for enhanced visualization.}  
    \label{fig.progressive}
\end{figure*} 

We denote our proposed LP-Net framework as $\Phi(\cdot)$, with the goal of recovering a dense, high-quality depth map $\hat{D}=\Phi(I, S)$ from sparse depth measurements $S$ and the corresponding color image $I$. The Laplacian Pyramid decomposes a depth map into multiple bandpass images and a low-resolution residual, capturing high-frequency details and the global scene structure, respectively. This decomposition forms the basis of our depth completion strategy, which inversely mimics the Laplacian Pyramid process to generate the final depth estimation from a global to local perspective. Our LP-Net is constructed on the U-Net \cite{unet} architecture, involving both encoding and decoding processes, and fully leverages decoder features at each scale to either predict the low-frequency residual or recover the high-frequency details.

The depth completion process within LP-Net is structured into five progressive steps, as shown in the right part of Fig. \ref{fig.framework}. We commence by predicting the low-frequency residual $\hat{D}^{(4)}$ at a scale of $\sfrac{1}{16}$, which captures the overall scene structure. This initial prediction is achieved through a Regression Head (RH) employing a convolutional layer to regress a coarse depth map $\hat{D}'^{(4)}$, followed by a confidence-based fusion with the downsampled sparse depth $\hat{S}^{(4)}$. Here, $\hat{S}^{(i)}$ for scale $i$ is derived through a weighted pooling operation akin to \cite{bp_net}. Specifically, with a downsampling factor of $s=2^i$, we divide the original sparse depth $S$ into non-overlapping $s \times s$ patches. We compute a positive weight $\omega^i$ for each pixel within a patch $P$ using a convolutional layer, followed by an exponential function. The sparse depth $\hat{S}^{(i)}$ is calculated by aggregating over the valid depth measurements within each patch:
\begin{equation}
    \hat{S}^{(i)}=\frac{\sum_{p\in P}\omega_p^iS_p}{\sum_{p\in P}\mathbb{I}(S_p)\omega_p^i + \epsilon},
\end{equation}
where $\mathbb{I}(\cdot)$ verifies the validity of pixel $S_p$, and $\epsilon$ is a small constant to prevent division by zero. We use a pixel-wise confidence $c$ to gauge the reliability of $\hat{S}^{(i)}$ and $S$. This confidence is estimated using a convolutional layer $\text{conv}(\cdot)$ and a sigmoid function $\sigma(\cdot)$:
\begin{equation}
    c_i=\sigma(\text{conv}(\text{cat}(F_d^i, \hat{S}^{(i)}))),
    \label{eq.ci}
\end{equation}
where $\text{cat}(\cdot)$ denote concatenation along the channel dimension. The final depth output $\hat{D}^{(4)}$ is obtained through a linear interpolation between $\hat{D}'^{(4)}$ and $S^{(4)}$ with confidence $c_4$, which can also be extended to any scale $i$:
\begin{equation}
    \hat{D}^{(i)}=c_i\cdot \hat{S}^{(i)}+(1-c_i)\cdot \hat{D}'^{(i)},
    \label{eq.fuse}
\end{equation}
where we use $\hat{D}^{(0)}=\hat{D}$ and $\hat{S}^{(0)}=S$ when fusing depth at the full resolution step.

Subsequently, the initial low-frequency residual $\hat{D}^{(4)}$ undergoes four further steps of consecutive upsampling and refinement to yield the final depth completion result $\hat{D}$. At each scale $i$, the prediction from the previous step $\hat{D}^{(i+1)}$ is upsampled by a factor of 2 using bilinear interpolation to produce the coarse estimation $\hat{D}'^{(i)}$. This is then fused with $\hat{S}^{(i)}$ using the confidence $c_i$ as previously described in Eq. \ref{eq.ci} and Eq. \ref{eq.fuse}. Instead of directly estimating the bandpass image at each scale as in traditional Laplacian Pyramid reconstruction, we employ a novel selective filtering approach through the Selective Depth Filtering (SDF) module. This module adaptively learns both a smoothness filter and a sharpness filter to enhance smooth transition areas and boundary details in the depth map, respectively. The intermediate results of this depth completion process are visually presented in Fig. \ref{fig.progressive}, providing insight into the progressive enhancement of the depth map from coarse to detailed predictions. This method not only reduces the complexity of the depth completion problem but also enhances computational efficiency by focusing on critical details at each resolution level.

\subsection{LP-Net Architecture}

\begin{figure}
    \centering
    \includegraphics[width=0.96\linewidth]{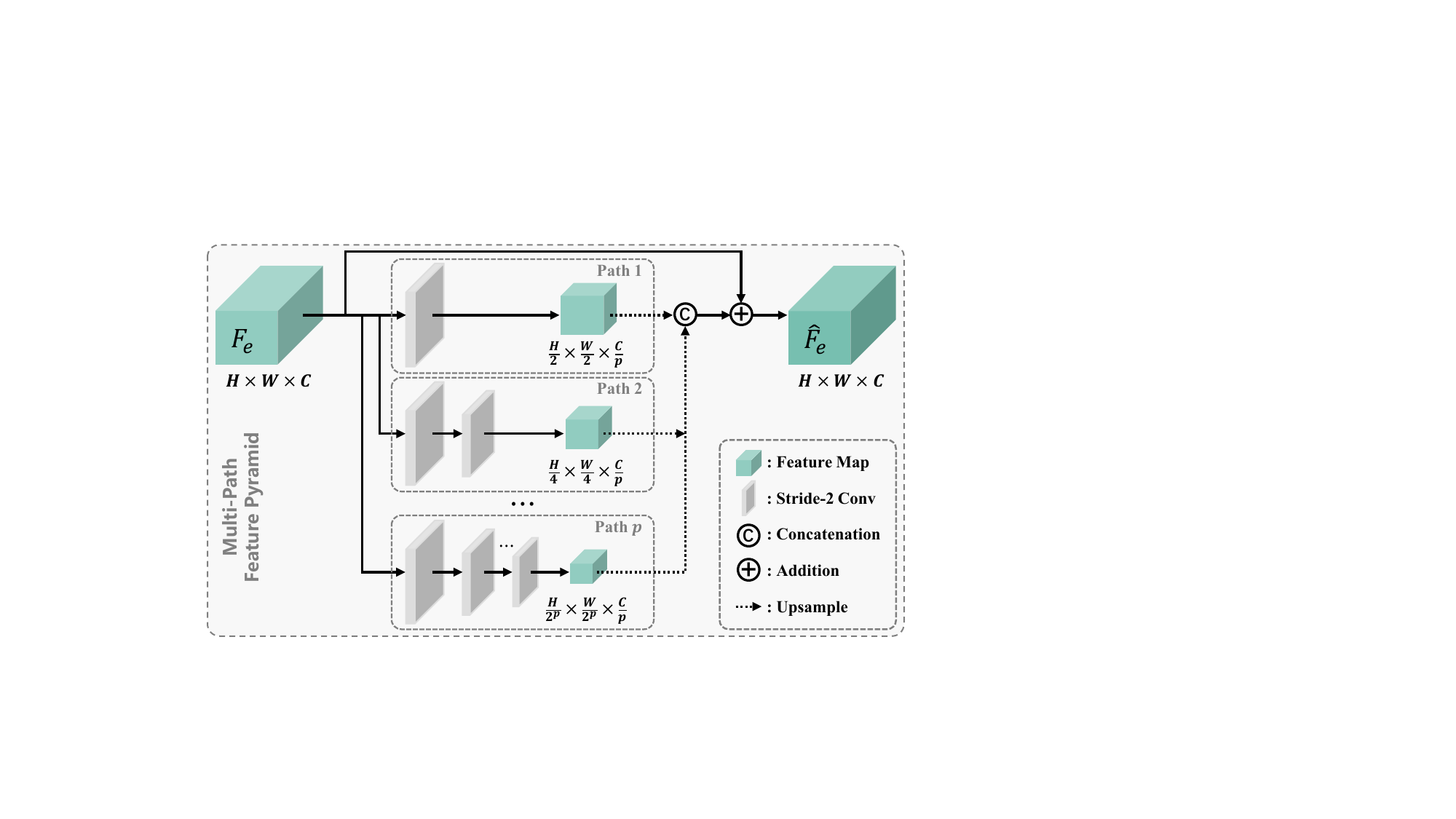}
    \caption{Illustration of the Multi-path Feature Pyramid (MFP) Module. This module segments the feature map $F_e$ into $p$ pathways, which are then transformed through multiple convolutional layers with a stride of 2. These pathways are subsequently upsampled and fused to integrate global information across different visual fields, yielding $\hat{F}_e$.}  
    \label{fig.mfp}
\end{figure} 

\subsubsection{Overall}

We visually depict the overall architecture of LP-Net in Fig. \ref{fig.framework}. LP-Net adopts a widely used U-Net \cite{unet} architecture, which includes an encoding process for extracting features from both the color image $I$ and the sparse depth $S$ modalities, followed by a decoding process that progressively reconstructs features at each scale. This reconstruction is enhanced by concatenating with encoder features at the corresponding scale. We utilize two separate encoders, $\text{E}_{img}(\cdot)$ for the color image and $\text{E}_{dep}(\cdot)$ for the sparse depth:
\begin{equation}
    \{F_{img}^i\}_{i=0}^4=\text{E}_{img}(I),\quad \{F_{dep}^i\}_{i=0}^4=\text{E}_{dep}(S).
\end{equation}
These encoders share an identical architecture with their respective parameters. They operate across five scales, ranging from $\sfrac{1}{1}$ to $\sfrac{1}{16}$. The input image is initially processed through a convolutional layer and then by a residual block \cite{resnet} to extract features at full resolution. Each of the subsequent four scales follows the same structure, comprising two residual blocks for downsampling and feature extraction. At the conclusion of each scale, features from both modalities are fused via concatenation, followed by a convolutional layer to reduce the feature dimensionality.

Upon feature fusion at the deepest scale, we employ a Multi-path Feature Pyramid (MFP) module to harness the global information embedded within the features, thereby bolstering the model's global perception capabilities. The decoding process unfolds over five steps. Initially, the deepest decoder feature $F_d^4$ is fed into a Regression Head module to generate the lowest-resolution depth map, \ie, the low-frequency residual. Then, $F_d^4$ undergoes four deconvolution operations to incrementally reconstruct the decoder features from $F_d^3$ to $F_d^0$. At each scale, the decoder feature is utilized by the Selective Depth Filtering (SDF) module to refine the high-frequency details of the previously predicted depth map. In the following sections, we will delve into the details of the MFP and SDF modules.

\subsubsection{Multi-path Feature Pyramid Module}

\begin{figure}
    \centering
    \includegraphics[width=\linewidth]{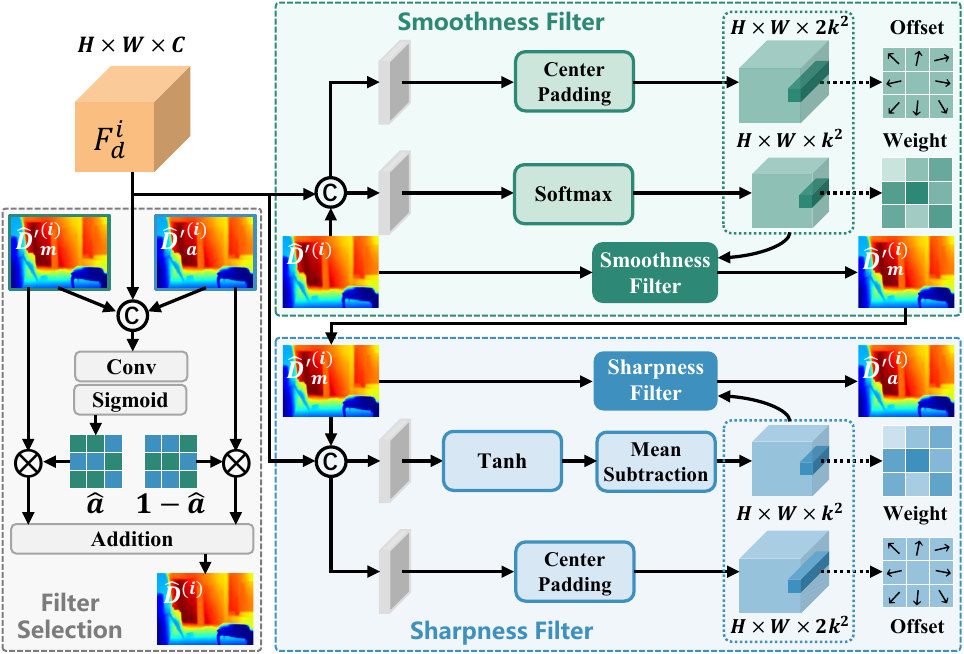}
    \caption{Illustration of the Selective Depth Filtering (SDF) Module. This module leverages the decoder feature $F_d^{i}$ to dynamically learn a smoothness filter (top) and a sharpness filter (bottom), which are sequentially applied to the input depth map $\hat{D}'^{(i)}$ at the $i$-th scale, producing $\hat{D}'^{(i)}_m$ and $\hat{D}'^{(i)}_a$, respectively. The final depth map $\hat{D}^{(i)}$ is then predicted via a selective mechanism. Here, $\otimes$ signifies element-wise multiplication, $k$ denotes the kernel size, and $\hat{a}$ represents the attention map.}  
    \label{fig.sdf}
\end{figure} 

The Multi-path Feature Pyramid (MFP) module is designed to integrate global information from various visual fields, thereby improving the model's global perception capabilities. Deployed to refine the deepest encoder feature $F_e^4$, MFP aids in the more precise estimation of the global scene structure $\hat{D}^{(4)}$. The detailed architecture of the MFP is depicted in Fig. \ref{fig.mfp}. MFP adheres to the split-transform-merge design paradigm. It initially splits the input feature map $F_e\in\mathbb{R}^{H \times W \times C}$ along the channel dimension, leading to $p$ feature maps each with the same resolution and $\frac{C}{p}$ channels. 
Each of these squeezed feature maps is then processed by a distinct sub-network, which includes several convolutional layers with a stride of 2. Specifically, the $i$-th sub-network contains $i$ convolutional layers, downsampling the input feature map by a factor of $2^i$. This configuration allows each pathway to capture global information from a specific visual field. Following this, these feature maps are upsampled to the original resolution of $H \times W$ using bilinear interpolation and then fused through concatenation, followed by a convolutional layer. The resulting fused feature map is added to the original input feature map $F_e$, yielding an enhanced output $\hat{F}_e$, which significantly boosts the model's global perception capabilities.

\subsubsection{Selective Depth Filtering Module}

The Selective Depth Filtering (SDF) module is introduced to refine the coarse depth map $\hat{D}'^{(i)}$ after its fusion with the sparse depth measurement $\hat{S}^{(i)}$ at scale $i$. The detailed architecture of the SDF module is illustrated in Fig. \ref{fig.sdf}. Diverging from previous approaches which focus solely on either a smoothness filter \cite{dyspn, cspn} or a sharpness filter \cite{lrru, dfu}, our SDF module simultaneously learns both types of filters and amalgamates their benefits through a selection mechanism. Specifically, the coarse depth $\hat{D}'^{(i)} \in \mathbb{R}^{H \times W}$ and the corresponding decoder feature $F_d^{i} \in \mathbb{R}^{H \times W \times C}$ are first concatenated and fed into two convolutional layers. These two layers are responsible for generating the weights $\omega \in \mathbb{R}^{H \times W \times k^2}$ and offsets $o \in \mathbb{R}^{H \times W \times 2k^2}$, respectively, for deformable depth filtering \cite{dcn, dcnv2}. 

For offset generation, we zero-pad the center element of each kernel, ensuring the original input depth map is included in the filtering operation. To generate weights for the smoothness kernel, we normalize the generated weights using a $\text{Softmax}(\cdot)$ function, ensuring they sum to $1$. The smoothness filter $f_{m}(\cdot)$ then employs these weights $\omega_m$ and offsets $o_m$ to process the depth map $\hat{D}'^{(i)}$:
\begin{equation}
    \hat{D}'^{(i)}_m = f_m(\hat{D}'^{(i)}, \omega_m, o_m).
\end{equation}
For the sharpness filter, we first constrain the kernel weights to the range $(-1, 1)$ using a $\text{tanh}(\cdot)$ function and then normalize by subtracting the mean of the kernel weights, ensuring they sum to $0$. The sharpness filter $f_a(\cdot)$ then applies these weights $\omega_a$ and offsets $o_a$ to the depth map $\hat{D}'^{(i)}_m$:
\begin{equation}
    \hat{D}'^{(i)}_a = f_a(\hat{D}'^{(i)}_m, \omega_a, o_a).
\end{equation}
Note that, unlike previous propagation-based methods that rely on iterative filtering, both our smoothness filter and sharpness filter are applied only once to the input depth map, thereby significantly reducing computational cost.

\begin{figure}
    \centering
    \includegraphics[width=0.96\linewidth]{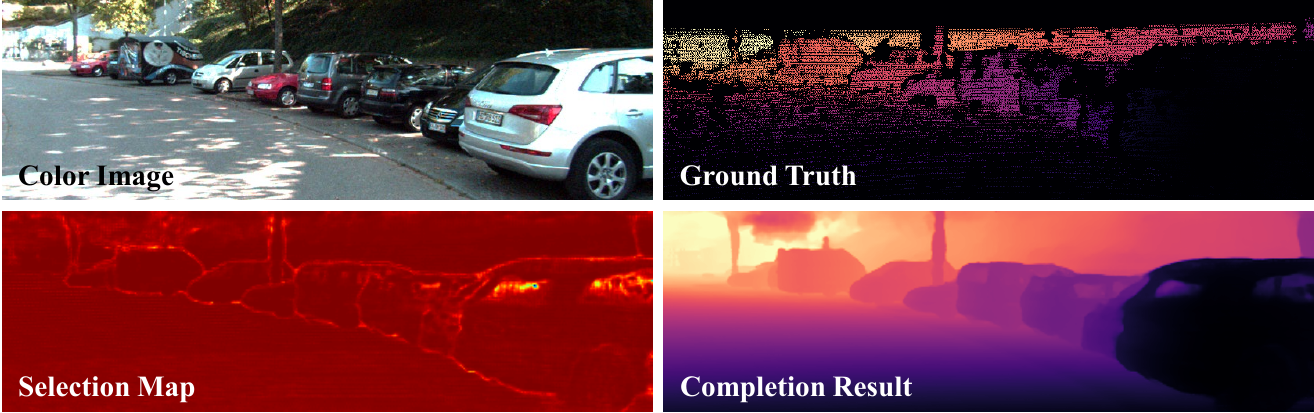}
    \caption{Illustration of the selective filtering mechanism in the SDF module. In the selection map, redder colors indicate a greater adoption of the smoothness filter, which is predominantly seen in the smoother areas of the depth map, whereas the sharpness filter is concentrated at the boundaries, demonstrating the efficacy of our proposed selective filtering mechanism.}  
    \label{fig.selective}
\end{figure} 

The smoothness filter reduces noise in the coarse prediction, enhancing the depth map by integrating confident sparse measurements from $\hat{D}'^{(i)}$, while the sharpness filter enhances depth boundaries, thus improving the details. We employ an attention-based selection mechanism to combine the strengths of both filters. Specifically, we concatenate the decoder feature $F_d^i$ with the intermediate depth maps $\hat{D}'^{(i)}_m$ and $\hat{D}'^{(i)}_a$, and generate a spatial attention map $\hat{a}$ by applying a convolution followed by a $\text{Sigmoid}(\cdot)$ function. The final depth map $\hat{D}^{(i)}$ is then computed as a linear interpolation between these two filtered depth predictions:
\begin{equation}
    \hat{D}^{(i)} = \hat{a} \cdot \hat{D}'^{(i)}_m + (1 - \hat{a}) \cdot \hat{D}'^{(i)}_a.
\end{equation}
We visually illustrate this selection mechanism in Fig. \ref{fig.selective}, where the selection map $\hat{a}$ (at the left bottom corner) shows higher values in smooth areas of the depth map and lower values at the boundaries, effectively leveraging the respective advantages of the two filters.

\subsection{Loss Function}

Our LP-Net is trained in an end-to-end manner under the supervision of ground truth depth maps $D$. We employ a combination of Mean Squared Error (MSE) and Mean Absolute Error (MAE) to penalize incorrect depth predictions, and introduce a multi-scale constraint to provide effective supervision at each scale $i$. The total training loss $L$ can be mathematically written as follows: 
\begin{equation}
    \begin{aligned}
        L=\sum_{i=0}^4\sum_{p\in\mathcal{V}} &\Vert\text{up}(\hat{D}^{(i)})_p - D_p\Vert_2^2+\\&\Vert\text{up}(\hat{D}^{(i)})_p - D_p\Vert_1,
    \end{aligned}
\end{equation} 
where $\mathcal{V}$ denotes the set of valid pixels in $D$, $\text{up}(\cdot)$ upsamples the depth predictions $\hat{D}^{(i)}$ to match the resolution of $D$ using bilinear interpolation, and $\hat{D}^{(0)}=\hat{D}$ represents the final depth completion result at full resolution.

\section{Experiment}

In this section, we substantiate the efficacy of LP-Net by benchmarking it against existing state-of-the-art (SOTA) approaches. We begin by describing the datasets utilized for evaluation and outlining the metrics employed. Subsequently, we detail the implementation specifics of our approach. We then present comparative results across three public benchmarks to illustrate the superiority of our method. Lastly, we validate the contributions of each component within LP-Net through comprehensive ablation studies.

\begin{table}
    \renewcommand\arraystretch{1.2}
    \setlength{\tabcolsep}{3.6pt}
    \caption{Quantitative comparison on KITTI DC online leaderboard. The best and second-best methods are highlighted in bold and underlined, respectively.}
    \label{tab.kitti}
    \centering
    \resizebox{\linewidth}{!}{
        \begin{tabular}{l||c||cccc}
            \Xhline{1.5pt}
            Method & Publication & \begin{tabular}[c]{@{}c@{}}RMSE $\downarrow$\\ (mm)\end{tabular} &
            \begin{tabular}[c]{@{}c@{}}MAE $\downarrow$\\ (mm)\end{tabular} &
            \begin{tabular}[c]{@{}c@{}}iRMSE $\downarrow$\\ (1/km)\end{tabular} &
            \begin{tabular}[c]{@{}c@{}}iMAE $\downarrow$\\ (1/km)\end{tabular}\\ \hline
            CSPN \cite{cspn}                & ECCV 2018  & 1019.64 & 279.46          & 2.93          & 1.15          \\
            S2D \cite{s2d}                   & ICRA 2019  & 814.73  & 249.95          & 2.80          & 1.21          \\
            DLiDAR \cite{deeplidar}  & CVPR 2019  & 758.38  & 226.50          & 2.56          & 1.15          \\
            FuseNet \cite{chen2019learning} & ICCV 2019  & 752.88                       & 221.19          & 2.34          & 1.14          \\
            NConv \cite{nconv}             & TPAMI 2020  & 829.98                       & 233.26          & 2.60          & 1.03          \\
            CSPN++ \cite{cspn++}           & AAAI 2020  & 743.69                       & 209.28          & 2.07          & 0.90          \\
            NLSPN \cite{nlspn}           & ECCV 2020  & 741.68                       & 199.59          & 1.99          & 0.84          \\
            GuideNet \cite{guidenet}        & TIP 2020   & 736.24                       & 218.83          & 2.25          & 0.99          \\
            TWISE \cite{twise}             & CVPR 2021  & 840.20                       & 195.58          & 2.08          & 0.82          \\
            ACMNet \cite{acmnet}          & TIP 2021   & 744.91                       & 206.09          & 2.08          & 0.90          \\
            FCFRNet \cite{fcfrnet}              & AAAI 2021  & 735.81                       & 217.15          & 2.20          & 0.98          \\
            PENet \cite{penet}                & ICRA 2021  & 730.08                       & 210.55          & 2.17          & 0.94          \\
            RigNet \cite{rignet}             & ECCV 2022  & 712.66                       & 203.25          & 2.08          & 0.90          \\
            CFormer \cite{cformer}              & CVPR 2023  & 708.87                       & 203.45          & 2.01          & 0.88          \\
            DGDF \cite{dgdf}                           & TCSVT 2023 & 707.93                       & 205.11          & 2.05          & 0.91          \\
            DySPN \cite{lin2023dyspn}             & TCSVT 2023 & 700.16                       & 189.70          & 1.84          & 0.81          \\
            BEV@DC \cite{bev_dc}                    & CVPR 2023  & 697.44                       & 189.44          & 1.83          & 0.82          \\
            LRRU \cite{lrru}                & ICCV 2023  & 696.51                       & 189.96          & 1.87          & 0.81          \\
            TPVD \cite{tpvd}                        & CVPR 2024  & 693.97                       & 188.60          & 1.82          & 0.81          \\
            DFU \cite{dfu}                           & CVPR 2024  & 686.46                       & {\ul 187.95}    & 1.83          & {\ul 0.81}    \\
            BP-Net \cite{bp_net}                    & CVPR 2024  & {\ul 684.90}                 & 194.69          & {\ul 1.82}    & 0.84          \\ 
            OGNI-DC \cite{ogni-dc} & ECCV 2024 & 708.38 & 193.20 & 1.86 & 0.83 \\ 
            \hline
            LP-Net (Our)                   & -- & \textbf{684.71}              & \textbf{186.63} & \textbf{1.81} & \textbf{0.80} \\ \hline
            \Xhline{1.5pt}
        \end{tabular}
    }
\end{table}

\subsection{Dataset}

We assess our method, LP-Net, on three public depth completion benchmarks, encompassing both outdoor and indoor environments. These datasets include the KITTI Depth Completion (DC) dataset \cite{kitti}, the NYUv2 dataset \cite{nyu}, and the TOFDC dataset \cite{tpvd}.

\textbf{KITTI DC} is a well-known outdoor dataset featuring color images from a stereo camera and ground truth depth from temporally registered LiDAR scans, further validated by stereo image pairs. It comprises over 80K training samples and 1,000 validation samples. Additionally, there are 1,000 testing samples available for evaluation on an online server with a public leaderboard. For training, we crop the bottom of the images from $1216\times 352$ to $1216\times 256$ resolution to exclude top areas without valid depth measurements.

\textbf{NYUv2} consists of indoor scenes captured by a Kinect sensor, totaling 464 scenes. We use 50K frames from 249 scenes for training as proposed by Ma \etal \cite{s2d}, with the official test set comprising 654 samples from 215 scenes for evaluation. Following established practices \cite{nlspn, rignet}, we downsample the images from $640\times 480$ to $320\times 240$ resolution, then center-crop them to $304\times 228$. Sparse depth for each sample is created by randomly selecting 500 points from the ground truth depth maps. Since our network requires an input resolution divisible by 16, we further pad the images to $304\times 240$ with border values. During validation, evaluations are confined to valid regions for a fair comparison with other methods.

\textbf{TOFDC} is an indoor dataset acquired with a mobile phone equipped with a Time-of-Flight (ToF) camera. The mobile phone records color images and low-resolution, incomplete depth maps, while high-precision ground truth depth maps are collected by an Helios ToF camera. This dataset features a variety of subjects including flowers, human figures, and toys, set in diverse scenes and lighting conditions. It includes 10K samples for training and 560 for evaluation. The sparse depth is generated by upsampling the low-resolution depth maps via nearest interpolation to align with the resolution of color images and ground truth depth maps at $512\times 384$ resolution.

\begin{figure*}
    \centering
    \includegraphics[width=0.98\linewidth]{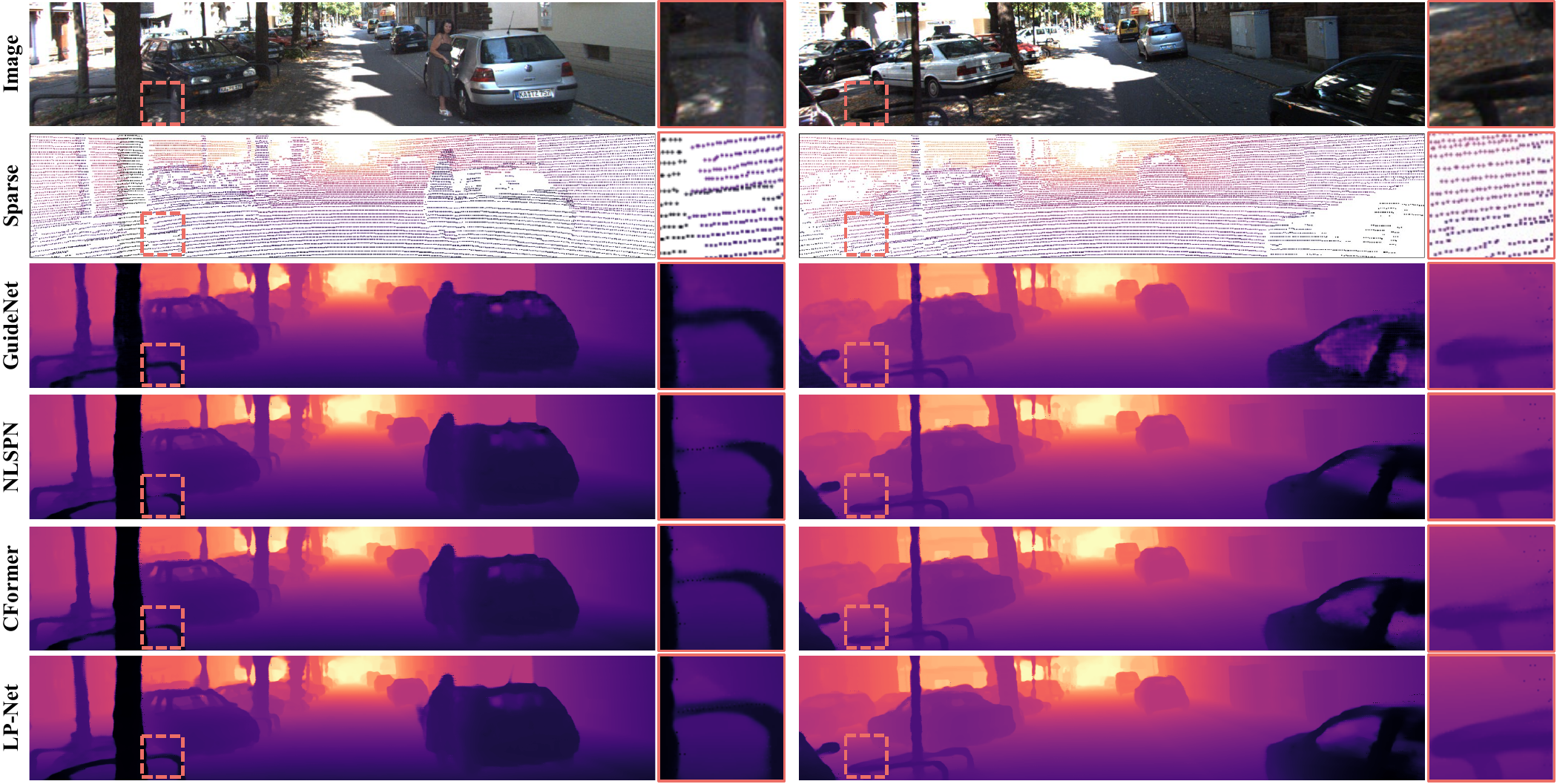}
    \caption{Qualitative comparison on the official KITTI DC test set. From top to bottom: color images, sparse depth maps, and results from GuideNet \cite{guidenet}, NLSPN \cite{nlspn}, CFormer \cite{cformer}, and our LP-Net method, respectively. Regions where our method outperforms the competitors are highlighted with red boxes, with detailed comparisons shown to the right of each image.}  
    \label{fig.compare_kitti}
\end{figure*} 

\subsection{Implementation Detail}

The LP-Net is implemented using the PyTorch \cite{pytorch} library and trained from scratch on four NVIDIA RTX 4090 GPUs using the distributed data parallel technique. We utilize the residual block \cite{resnet} as the foundational building block and incorporate the DropPath \cite{droppath} technique at the end of each residual block to prevent overfitting. The drop ratio is linearly increased from 0 to 0.5 across all residual blocks in the encoder. We train LP-Net for roughly 300K, 200K, and 25K iterations on the KITTI DC, NYUv2, and TOFDC datasets, respectively, with a total batch size of 8. The AdamW optimizer \cite{adamw} is employed along with the OneCycle learning rate scheduler \cite{onecycle}. We employ a weight decay of $0$, $10^{-5}$ and $10^{-4}$ for the KITTI DC, NYUv2, and TOFDC datasets, respectively, to regularize the training process. The learning rate scheduler involves an initial increase from $2 \times 10^{-5}$ to $10^{-3}$ over the first $10\%$ of iterations, followed by a decrease to $2 \times 10^{-4}$ using a cosine annealing strategy. We randomly perturb the learning rate by up to $10\%$ at each iteration for better model convergence. To improve generalization and combat overfitting, we incorporate various data augmentation techniques in our training pipeline, including random horizontal flips, random rotations, random crop and random color jitter. The final model is derived using the Exponential Moving Average (EMA) technique with a decay rate of 0.9999.

\subsection{Evaluation Metric}

We employ eight metrics to assess the depth completion performance of each model. These include Root Mean Square Error (RMSE), Mean Absolute Error (MAE), inverse RMSE (iRMSE), inverse MAE (iMAE), Absolute Relative Error (REL) and Threshold Accuracy ($\sigma_{1}, \sigma_2, \sigma_3$). Let $\hat{D}$ and $D$ denote the predicted and ground truth depth, respectively. These metrics can be mathematically defined as:
\begin{equation}
    \begin{aligned}
        &\text{RMSE}\ (mm):\sqrt{\frac{1}{\vert\mathcal{V}\vert}\sum_{p\in\mathcal{V}}(\hat{D}_p-D_p)^2}\\
        &\text{MAE}\ (mm):\frac{1}{\vert\mathcal{V}\vert}\sum_{p\in\mathcal{V}}\vert\hat{D}_p-D_p\vert\\
        &\text{iRMSE}\ (1/km):\sqrt{\frac{1}{\vert\mathcal{V}\vert}\sum_{p\in\mathcal{V}}(\frac{1}{\hat{D}_p}-\frac{1}{D_p})^2}\\
        &\text{iMAE}\ (1/km):\frac{1}{\vert\mathcal{V}\vert}\sum_{p\in\mathcal{V}}\vert\frac{1}{\hat{D}_p}-\frac{1}{D_p}\vert\\
        &\text{REL}:\frac{1}{\vert\mathcal{V}\vert}\sum_{p\in\mathcal{V}}\vert\frac{\hat{D}_p-D_p}{D_p}\vert\\
        &\sigma_{i=1,2,3}\ (\%):max(\frac{\hat{D}_p}{D_p},\frac{D_p}{\hat{D}_p})<1.25^i,
    \end{aligned}
\end{equation}
where $\mathcal{V}$ represents the set of valid pixels in $D$, and $\vert\mathcal{V}\vert$ denotes the cardinality of $\mathcal{V}$. The metric units appear to the right of each metric name, where the REL denotes a ratio.

\subsection{Comparison with SOTA Methods}

\begin{figure*}
    \centering
    \includegraphics[width=0.98\linewidth]{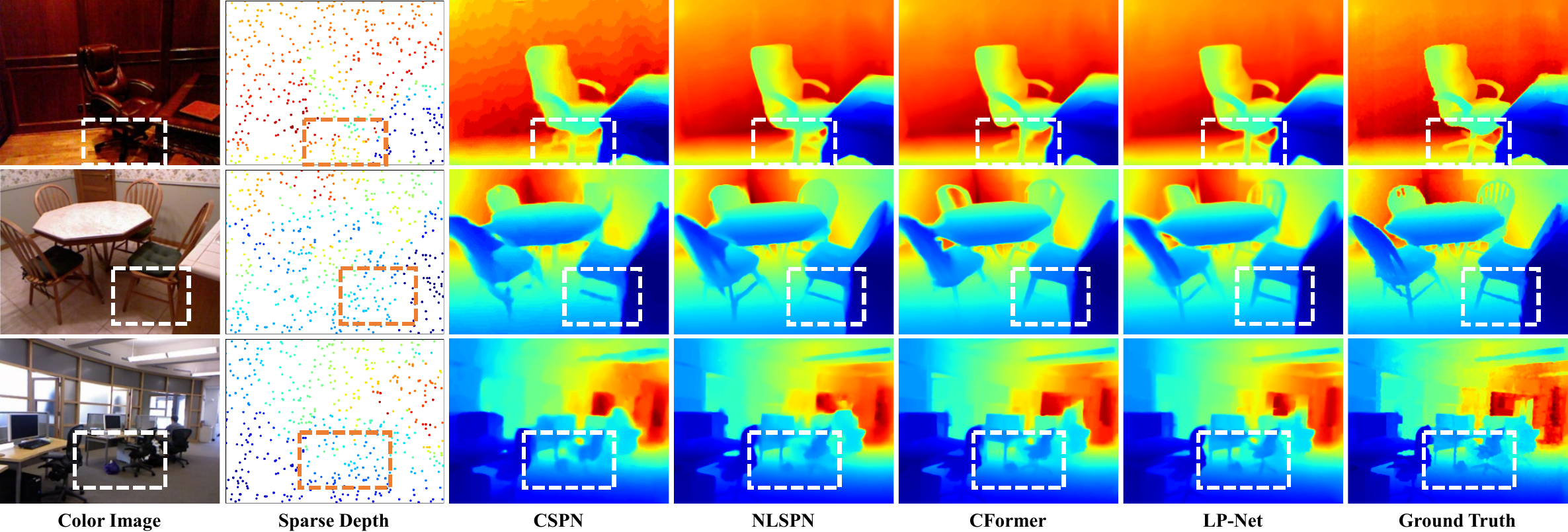}
    \caption{Qualitative comparison on the NYUv2 test set. From left to right: color images, sparse depth maps, results from CSPN \cite{cspn}, NLSPN \cite{nlspn}, CFormer \cite{cformer}, and our LP-Net method, respectively, and ground truth depth maps. Regions where our method outperforms the competitors are highlighted with boxes.}  
    \label{fig.compare_nyu}
    \vskip -0.1in
\end{figure*} 

\begin{table}
    \renewcommand\arraystretch{1.2}
    \setlength{\tabcolsep}{3.6pt}
    \caption{Quantitative comparison on NYUv2 dataset.}
    \label{tab.nyu}
    \centering
    \resizebox{\linewidth}{!}{
        \begin{tabular}{l||c||ccc}
            \Xhline{1.5pt}
            Method & Publication & RMSE $\downarrow$ (m) & REL $\downarrow$ & ${\delta }_{1}$ $\uparrow$ (\%) \\ \hline
            CSPN \cite{cspn}           & ECCV 2018  & 0.117          & 0.016          & 99.2          \\
            DLiDAR \cite{deeplidar}    & CVPR 2019  & 0.115          & 0.022          & 99.3          \\
            GuideNet \cite{guidenet}   & TIP 2020   & 0.101          & 0.015          & 99.5          \\
            NLSPN \cite{nlspn}         & ECCV 2020  & 0.092          & 0.012          & \textbf{99.6} \\
            FCFRNet \cite{fcfrnet}     & AAAI 2021  & 0.106          & 0.015          & 99.5          \\
            ACMNet \cite{acmnet}       & TIP 2021   & 0.105          & 0.015          & 99.4          \\
            RigNet \cite{rignet}       & ECCV 2022  & 0.090          & 0.013          & \textbf{99.6} \\
            GraphCSPN \cite{graphcspn} & ECCV 2022  & 0.090          & 0.012          & \textbf{99.6} \\
            DGDF \cite{dgdf}           & TCSVT 2023 & 0.098          & 0.014          & 99.5          \\
            DySPN \cite{lin2023dyspn}        & TCSVT 2023  & \textbf{0.089}        & 0.012            & \textbf{99.6}                      \\
            CFormer \cite{cformer}     & CVPR 2023  & 0.091          & 0.012          & \textbf{99.6} \\
            LRRU \cite{lrru}           & ICCV 2023  & 0.091          & \textbf{0.011} & \textbf{99.6} \\
            BEV@DC \cite{bev_dc}       & CVPR 2023  & \textbf{0.089} & 0.012          & \textbf{99.6} \\
            PointDC \cite{yu2023aggregating} & ICCV 2023   & \textbf{0.089}        & 0.012            & \textbf{99.6}                      \\
            DFU \cite{dfu}             & CVPR 2024  & 0.091          & \textbf{0.011} & \textbf{99.6} \\
            BP-Net \cite{bp_net}       & CVPR 2024  & \textbf{0.089} & 0.012          & \textbf{99.6} \\ \hline
            LP-Net (Our)               & -- & {\ul 0.090}    & {\ul 0.012}    & \textbf{99.6} \\ 
            \Xhline{1.5pt}
        \end{tabular}
    } 
\end{table}

We benchmark LP-Net against existing state-of-the-art (SOTA) methods on the KITTI DC, NYUv2, and TOFDC datasets to demonstrate its effectiveness. Additionally, we compare LP-Net with the latest approaches in terms of computational cost to illustrate its superior computational efficiency.

\subsubsection{KITTI DC}
Quantitative results on the official KITTI DC test set are presented in Tab. \ref{tab.kitti}, with all metrics sourced from the online KITTI DC leaderboard. Our method demonstrates superior performance compared to existing SOTA approaches. Unlike previous methods which excel in only certain metrics, LP-Net significantly improves across all four metrics. Specifically, it surpasses BP-Net, the second-best method, by $4.14\%$, $0.55\%$, and $4.76\%$ in MAE, iRMSE, and iMAE, respectively, evidencing the exceptional performance of our approach. Qualitative results are shown in Fig. \ref{fig.compare_kitti}, where regions in which LP-Net outperforms are highlighted with red boxes, and detailed comparisons are provided on the right of each image. According to these comparisons, LP-Net excels in recovering the correct local structures. For instance, LP-Net accurately estimates the structure of a ground-level railing, which other methods either omit or predict coarsely.

\subsubsection{NYUv2}
We present the quantitative results for the NYUv2 dataset in Tab. \ref{tab.nyu}. Our method achieves the second-best performance in RMSE and REL metrics and the best in the $\sigma_{1}$ metric, showcasing SOTA overall performance on this dataset. The qualitative comparison is illustrated in Fig. \ref{fig.compare_nyu}. Compared to competitors, our method more accurately recovers local structures, as highlighted by the boxes. Specifically, in the first two samples, LP-Net correctly predicts the detailed structure of chair legs, whereas other methods either miss these elements or produce blurry predictions. In the last sample, our method successfully captures the structures of both table and chair legs, while competitors fail to represent these accurately.

\begin{table}
    \renewcommand\arraystretch{1.2}
    \setlength{\tabcolsep}{3.2pt}
    \caption{Quantitative comparison on TOFDC dataset.}
    \label{tab.tofdc}
    \centering
    \resizebox{\linewidth}{!}{
        \begin{tabular}{l||ccccc}
            \Xhline{1.5pt}
            Method                & RMSE $\downarrow$ (m) & REL $\downarrow$ & ${\delta }_{1}$ $\uparrow$ (\%) & ${\delta }_{2}$ $\uparrow$ (\%) & ${\delta }_{3}$ $\uparrow$ (\%) \\ \hline
            CSPN \cite{cspn}          & 0.224       & 0.042          & 94.5          & 95.3       & 96.5          \\
            FusionNet \cite{van2019sparse}  & 0.116       & 0.024          & 98.3          & 99.4       & 99.7          \\
            GuideNet \cite{guidenet}  & 0.146       & 0.030          & 97.6          & 98.9       & 99.5          \\
            ENet \cite{penet}           & 0.231       & 0.061          & 94.3          & 95.2       & 97.4          \\
            PENet \cite{penet}          & 0.241       & 0.043          & 94.6          & 95.3       & 95.5          \\
            NLSPN \cite{nlspn}     & 0.174       & 0.029          & 96.4          & 97.9       & 98.9          \\
            GraphCSPN \cite{graphcspn} & 0.253       & 0.052          & 92.0          & 96.9       & 98.7          \\
            RigNet \cite{rignet}       & 0.133       & 0.025          & 97.6          & 99.1       & 99.7          \\
            DySPN \cite{dyspn}                    & 0.123      & 0.034          & 98.2          & 99.3       & {\ul 99.8}    \\
            CFormer \cite{cformer}        & 0.113       & 0.029          & \textbf{99.1} & 99.6       & \textbf{99.9} \\
            PointDC \cite{yu2023aggregating}  & 0.109       & 0.021          & {\ul 98.5}    & 99.2       & 99.6          \\
            TPVD \cite{tpvd}                    & {\ul 0.092} & \textbf{0.014} & \textbf{99.1} & {\ul 99.6} & \textbf{99.9} \\ \hline
            LP-Net (Our) & \textbf{0.081}        & {\ul 0.017}      & \textbf{99.1}                   & \textbf{99.7}                   & \textbf{99.9}                   \\ 
            \Xhline{1.5pt}
        \end{tabular}
    } 
\end{table}

\begin{figure}
    \centering
    \includegraphics[width=0.98\linewidth]{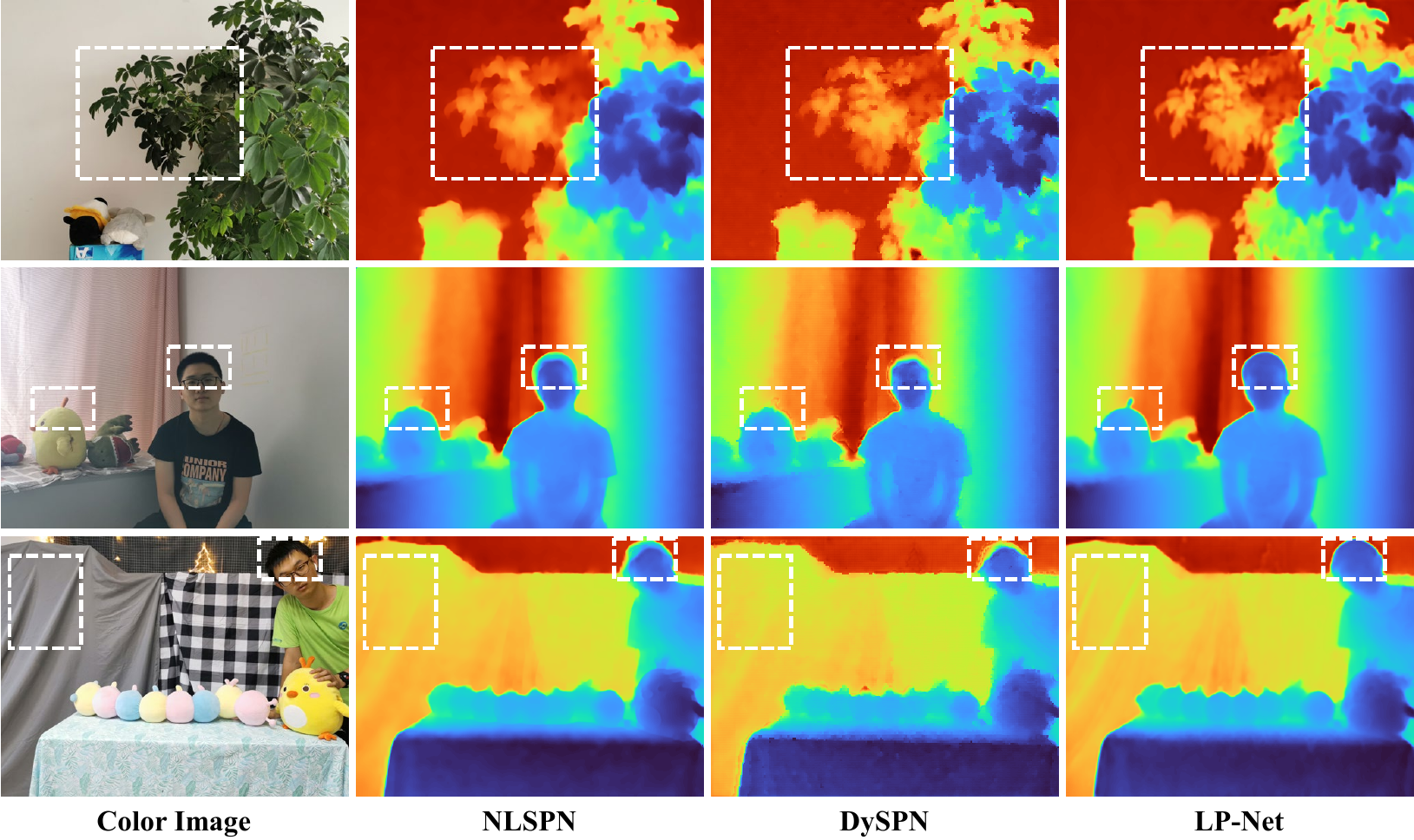}
    \caption{Qualitative comparison on the TOFDC test set. From left to right: color images, results from NLSPN \cite{nlspn}, DySPN \cite{dyspn} and our LP-Net method, respectively. Regions where our method outperforms the competitors are highlighted with white boxes.}  
    \label{fig.compare_tofdc}
    \vskip -0.05in
\end{figure} 

\subsubsection{TOFDC}
Quantitative results for the TOFDC dataset are presented in Tab. \ref{tab.tofdc}. Our method achieves the best performance on the RMSE, $\sigma_1$, $\sigma_2$, and $\sigma_3$ metrics while securing the second-best performance on the REL metric. Specifically, it improves the RMSE metric by $11.96\%$ compared to the second-best method, TPVD, underscoring the superior depth completion accuracy of our approach. The qualitative comparison is depicted in Fig. \ref{fig.compare_tofdc}. Our method yields accurate predictions with fine-grained details when compared to competitors. For instance, in the last two samples, LP-Net correctly estimates the structure of the human heads, whereas other approaches fail to capture the correct geometry. In the first and last samples, our method provides more detailed predictions for leaves and curtains, respectively, as highlighted by the white boxes in the figure.

\begin{table}
    \renewcommand\arraystretch{1.2}
    \setlength{\tabcolsep}{4pt}
    \caption{Quantitative result on computational efficiency. All methods are tested on $1216\times 256$ images using single RTX 4090 GPU.}
    \label{tab.cost}
    \centering
    \resizebox{\linewidth}{!}{
        \begin{tabular}{l||c||ccc|cc}
            \Xhline{1.5pt}
            Method &
            Prop &
            \begin{tabular}[c]{@{}c@{}}Param $\downarrow$\\ (M)\end{tabular} &
            \begin{tabular}[c]{@{}c@{}}Time $\downarrow$\\ (ms)\end{tabular} &
            \begin{tabular}[c]{@{}c@{}}Mem $\downarrow$\\ (GB)\end{tabular} &
            \begin{tabular}[c]{@{}c@{}}RMSE $\downarrow$\\ (mm)\end{tabular} &
            \begin{tabular}[c]{@{}c@{}}MAE $\downarrow$\\ (mm)\end{tabular} \\ \hline
            CFormer \cite{cformer} & $\checkmark$ & 83.5          & 78.30          & {\ul 1.96}    & 708.87          & 203.45          \\
            LRRU \cite{lrru}       & $\checkmark$ & \textbf{20.8} & 74.97          & 2.11          & 696.51          & 189.96          \\
            TPVD \cite{tpvd}       & $\checkmark$ & 31.2          & {\ul 74.27}    & 3.05          & 693.97          & 188.60          \\
            DFU \cite{dfu}         & $\checkmark$ & {\ul 25.6}    & 91.91          & 2.12          & 686.46          & {\ul 187.95}    \\
            BP-Net \cite{bp_net}   & $\checkmark$ & 89.9          & 83.59          & 6.19          & {\ul 684.90}    & 194.69          \\
            OGNI-DC \cite{ogni-dc} & $\checkmark$ & 84.4          & 123.49         & 1.97          & 708.38          & 193.20          \\ \hline
            LP-Net (our)           &              & 29.6          & \textbf{63.88} & \textbf{1.76} & \textbf{684.71} & \textbf{186.63} \\
            \Xhline{1.5pt}
        \end{tabular}
    } 
\end{table}

\subsubsection{Computational Efficiency}
To substantiate the computational efficiency of LP-Net, we compare its model size, inference speed, and memory usage with those of the most recent approaches. The quantitative results on the KITTI DC dataset are presented in Tab. \ref{tab.cost}, where "Prop" indicates the employment of propagation-based methods, while "Param", "Time", and "Mem" represent the number of network parameters, inference time per sample, and GPU memory usage, respectively. Our method not only achieves superior depth completion performance but also boasts the fastest inference speed and the lowest memory usage. Specifically, LP-Net surpasses BP-Net, the second-best method, by $23.58\%$ in inference speed and by $71.57\%$ in memory efficiency. This highlights the advantage of our multi-scale Laplacian Pyramid-based prediction scheme over the propagation-based methods that currently dominate the SOTAs.

\subsection{Ablation Study}

In this section, we illustrate the effectiveness of each component within LP-Net through comprehensive ablation studies. We initially conduct ablation experiments on the Multi-path Feature Pyramid (MFP) module and the Selective Depth Filtering (SDF) module. Subsequently, we analyze the robustness of our method across varying levels of depth sparsity. Finally, we highlight a distinctive feature of our approach, which adeptly balances prediction accuracy with inference speed.

\subsubsection{MFP Module} 
The MFP module is designed to integrate global information from various visual fields, thereby enhancing the estimation of the overall scene structure. Its effectiveness is evaluated through ablation studies as reported in Tab. \ref{tab.mfp}. The baseline model, labeled Variant-$\romannumeral1$, omits the MFP module from LP-Net, resulting in a significant performance drop of $0.91\%$ , $0.70\%$, $1.28\%$, and $0.51\%$ on RMSE, MAE, iRMSE, and iMAE metrics, respectively. Variant-$\romannumeral2$ substitutes the MFP module with a Vision Transformer (ViT) block \cite{vit}, which captures global information through self-attention. Although this variant markedly improves upon the baseline, it does not match the performance of the full LP-Net method, underscoring the necessity of the MFP module for integrating diverse global visual information. Additionally, we conducted further ablation on the MFP module to determine the optimal number of paths, with results shown in the middle section of the table. The inclusion of the MFP module consistently outperforms the baseline method, with the configuration using 4 paths yielding the best results.

\begin{table}
    \renewcommand\arraystretch{1.2}
    \setlength{\tabcolsep}{3.6pt}
    \caption{Ablation study of MFP module on KITTI DC validation set.}
    \label{tab.mfp}
    \centering
    \resizebox{\linewidth}{!}{
        \begin{tabular}{c||cc|c||cccc}
            \Xhline{1.5pt}
            Variant &
            \begin{tabular}[c]{@{}c@{}}MFP\\ Module\end{tabular} &
            \begin{tabular}[c]{@{}c@{}}ViT\\ Block\end{tabular} &
            \begin{tabular}[c]{@{}c@{}}MFP\\ Paths\end{tabular} &
            \begin{tabular}[c]{@{}c@{}}RMSE $\downarrow$\\ (mm)\end{tabular} &
            \begin{tabular}[c]{@{}c@{}}MAE $\downarrow$\\ (mm)\end{tabular} &
            \begin{tabular}[c]{@{}c@{}}iRMSE $\downarrow$\\ (1/km)\end{tabular} &
            \begin{tabular}[c]{@{}c@{}}iMAE $\downarrow$\\ (1/km)\end{tabular} \\ \hline
            $\romannumeral1$ &              &              & -- & 716.31          & 186.68          & 1.813          & 0.781          \\
            $\romannumeral2$ &              & $\checkmark$ & -- & {\ul 712.54}    & {\ul 186.01}    & {\ul 1.801}    & 0.783          \\ \hline
            $\romannumeral3$ & $\checkmark$ &              & 1 & 714.42          & 187.10          & 1.812          & 0.784          \\
            $\romannumeral4$ & $\checkmark$ &              & 2 & 714.10          & 186.36          & 1.805          & {\ul 0.780}    \\
            $\romannumeral5$ & $\checkmark$ &              & 3 & 715.63          & 186.41          & 1.804          & 0.781          \\ \hline
            LP-Net            & $\checkmark$ &              & 4 & \textbf{709.82} & \textbf{185.38} & \textbf{1.790} & \textbf{0.777} \\
            \Xhline{1.5pt}
        \end{tabular}
    }
    \vskip -0.05in
\end{table}

\begin{table}
    \renewcommand\arraystretch{1.2}
    \setlength{\tabcolsep}{3.6pt}
    \caption{Ablation study of SDF module on KITTI DC validation set. "Sel" and "Def" mean selective and deformable, respectively.}
    \label{tab.sdf}
    \centering
    \resizebox{\linewidth}{!}{
        \begin{tabular}{c||c||cccc||ccc}
            \Xhline{1.5pt}
            Variant &
            \begin{tabular}[c]{@{}c@{}}SDF\\ Module\end{tabular} &
            $f_m(\cdot)$ &
            $f_a(\cdot)$ &
            Sel &
            Def &
            \begin{tabular}[c]{@{}c@{}}RMSE $\downarrow$\\ (mm)\end{tabular} &
            \begin{tabular}[c]{@{}c@{}}MAE $\downarrow$\\ (mm)\end{tabular} &
            \begin{tabular}[c]{@{}c@{}}iMAE $\downarrow$\\ (1/km)\end{tabular} \\ \hline
            $\romannumeral 1$ &              &              &              &              &              & 717.23          & 194.85          & 0.822          \\ \hline
            $\romannumeral 2$ & $\checkmark$ & $\checkmark$ &              &              &              & {\ul 713.11}    & 186.24          & 0.779          \\
            $\romannumeral 3$ & $\checkmark$ &              & $\checkmark$ &              &              & 721,83          & 188.01          & 0.785          \\
            $\romannumeral 4$ & $\checkmark$ & $\checkmark$ & $\checkmark$ &              & $\checkmark$ & 714.19          & {\ul 186.10}    & {\ul 0.778}    \\
            $\romannumeral 5$ & $\checkmark$ & $\checkmark$ & $\checkmark$ & $\checkmark$ &              & 717.13          & 186.69          & 0.781          \\ \hline
            LP-Net            & $\checkmark$ & $\checkmark$ & $\checkmark$ & $\checkmark$ & $\checkmark$ & \textbf{709.82} & \textbf{185.38} & \textbf{0.777} \\ 
            \Xhline{1.5pt}
        \end{tabular}
    }
\end{table}

\subsubsection{SDF Module}

The SDF module is designed to recover high-frequency details in depth maps. We demonstrate its effectiveness through ablation studies presented in Tab. \ref{tab.sdf}. Variant-$\romannumeral 1$ follows the traditional approach of the Laplacian Pyramid, where a residual depth is predicted to refine the coarse depth prediction using a convolutional head. Compared to the full method, this variant shows a significant performance decrease of $1.04\%$, $5.11\%$, and $5.79\%$ on RMSE, MAE, and iMAE metrics, respectively, underscoring the superior effectiveness of our depth filtering mechanism for depth refinement. Next, we assess the individual contributions of the smoothness filter $f_m(\cdot)$ and the sharpness filter $f_a(\cdot)$ by applying each separately. The comparison between Variant-$\romannumeral 2$ and Variant-$\romannumeral 3$ reveals the better performance of the smoothness filter when used alone. Variant-$\romannumeral 4$ applies both $f_m(\cdot)$ and $f_a(\cdot)$  sequentially without the selection mechanism, resulting in a performance drop of $0.62\%$, $0.39\%$, and $0.13\%$ on the three metrics compared to the full method, which validates the effectiveness of our selective filtering mechanism. Lastly, Variant-$\romannumeral 5$ employs fixed neighbors for depth filtering and yields inferior results compared to the full method, highlighting the advantage of our deformable depth filtering approach.

\subsubsection{Robustness to Depth Sparsity}

\begin{figure}
    \centering
    \includegraphics[width=0.835\linewidth]{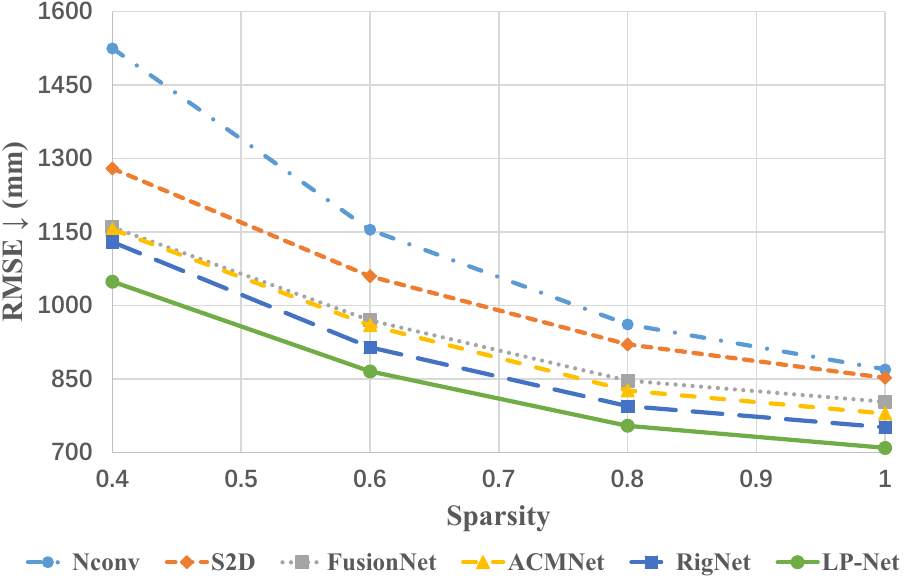}
    \caption{Quantitative comparison on the KITTI DC validation set. The original sparse depth input is randomly sampled to densities of 0.4, 0.6, and 0.8 to assess the robustness of methods across different levels of depth sparsity.}  
    \label{fig.sparsity}
    \vskip -0.05in
\end{figure} 

We compare our method with existing approaches in terms of robustness to input depth sparsity. The quantitative results are presented in Fig. \ref{fig.sparsity}. Following previous works \cite{guidenet, tpvd}, we generate input sparse depth at specific sparsity levels by uniformly sampling $40\%$, $60\%$, and $80\%$ of valid pixels from the original sparse depth, where the original sparse depth maps correspond to a sparsity of 1. As sparsity increases, all methods exhibit a significant performance decline; however, our method consistently outperforms competitors at each sparsity level, demonstrating superior robustness across various input depth sparsities.

\subsubsection{Accuracy vs. Speed} 

\begin{figure}
    \centering
    \includegraphics[width=0.85\linewidth]{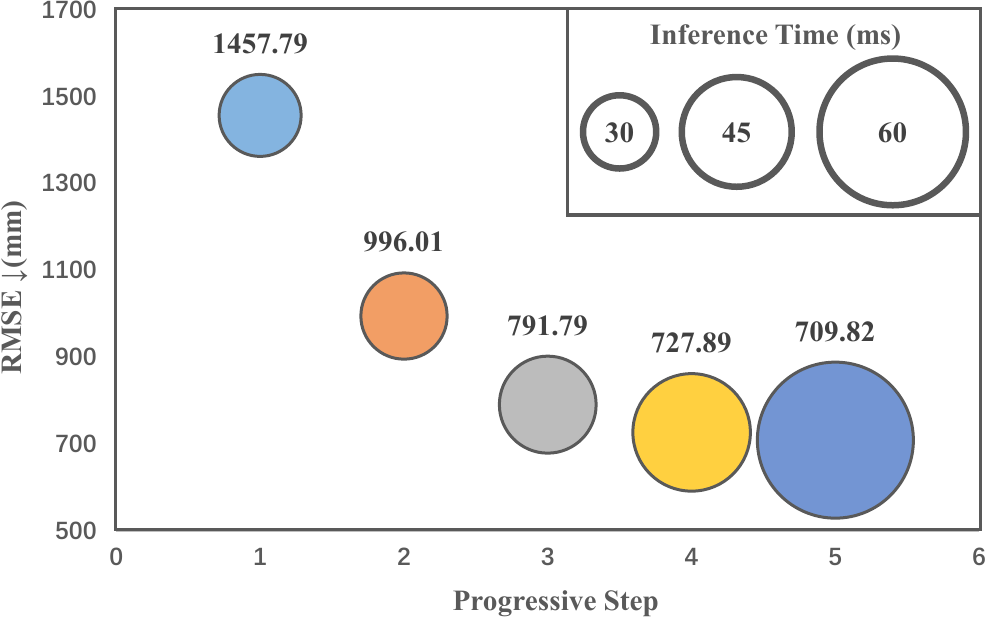}
    \caption{Relationship between depth completion accuracy and inference time at different progressive steps. The RMSE metric is reported at the top of each bubble, with the bubble's width representing the inference time per sample.}  
    \label{fig.avs}
    \vskip -0.05in
\end{figure}

LP-Net initially estimates the overall scene structure at $\sfrac{1}{16}$ of the output resolution. Subsequently, the depth maps are progressively upsampled while adding high-frequency details at each scale to recover local structures. This multi-scale, progressive prediction strategy enables us to balance depth completion accuracy with inference speed effectively by adjusting the number of progressive steps used for the final depth output. We visually depict the relationship between depth completion accuracy and inference time at each progressive step in Fig. \ref{fig.avs}. The depth maps predicted at lower resolutions are upsampled using bilinear interpolation to match the input resolution. Given the bottom-to-top progressive prediction approach of our method, the initial steps yield significant accuracy improvements with only a slight increase in inference time. Therefore, for applications where computational efficiency is paramount, fewer progressive steps can be utilized to achieve faster inference speeds. Conversely, full-step prediction should be employed for applications where high prediction accuracy is critical.

\section{Limitation and Future Work}

Developing a universal depth completion model capable of consistently generating high-quality, dense depth maps from any scene structure and sparse depth distribution remains a compelling research direction in the field of depth completion. Like most existing works, our method is currently tailored for depth completion within specific scene types and sparse depth distributions, which confines its applicability across a wide range of scenarios. Therefore, our future work will concentrate on evolving LP-Net into a universal model to enhance its prediction accuracy and practical utility.

\section{Conclusion}

In this paper, we introduce LP-Net, a novel depth completion framework that employs a multi-scale, progressive prediction scheme grounded in Laplacian Pyramid decomposition. We propose the Multi-path Feature Pyramid (MFP) module to integrate global information across various visual fields and the Selective Depth Filtering (SDF) module to recover high-frequency local structures at each scale. Through extensive experiments, we have demonstrated that LP-Net achieves superior depth completion accuracy and computational efficiency, while also highlighting its practical ability to balance prediction accuracy with inference speed.

\bibliographystyle{IEEEtran}
\bibliography{references}

\vskip -0.32in
\begin{IEEEbiography}
    [{\includegraphics[width=1in,height=1.25in,clip,keepaspectratio]{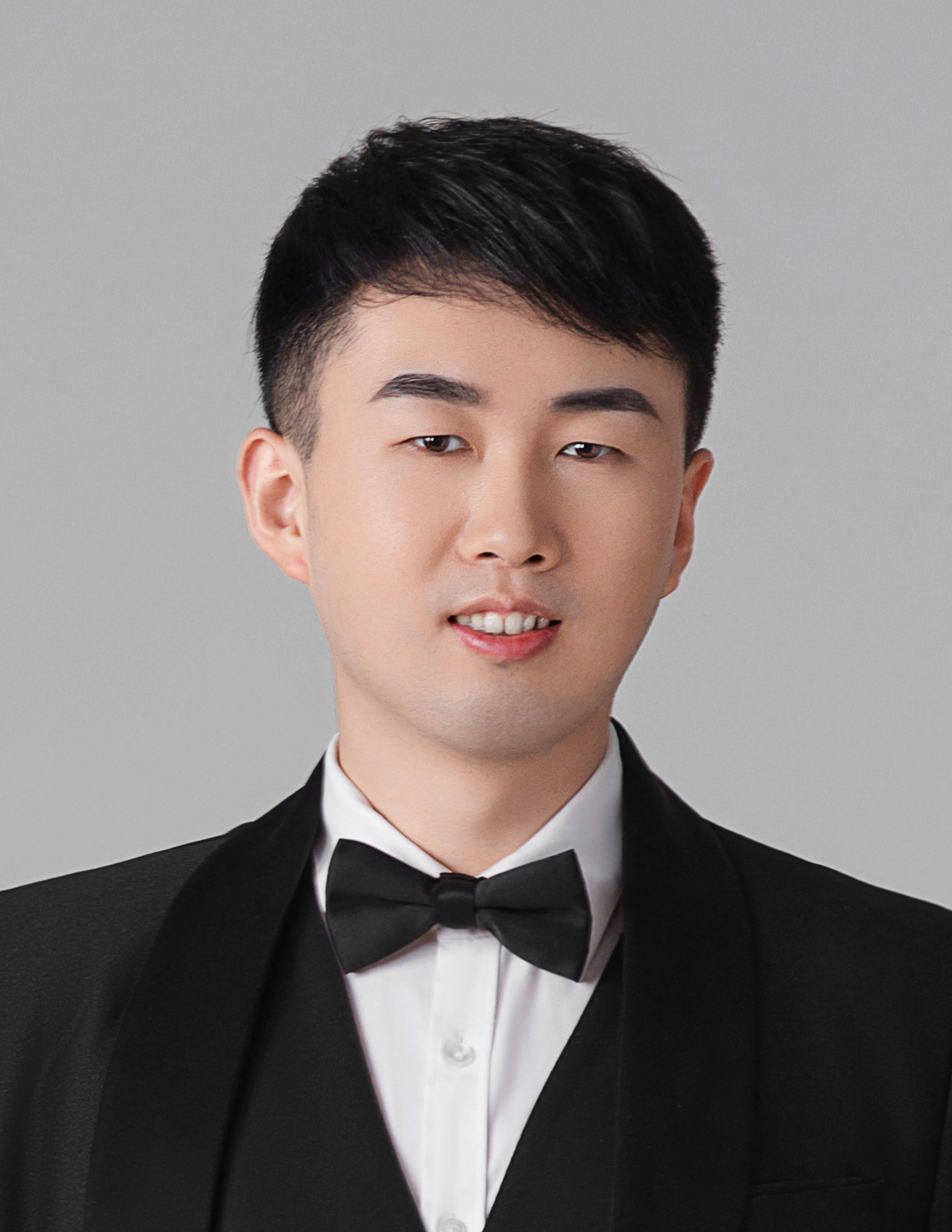}}]{Kun Wang} received his B.E. degree in Smart Grid Information Engineering from Qingdao University of Science and Technology in 2017. He is currently pursuing a Ph.D. in Control Science and Engineering at Nanjing University of Science and Technology under the supervision of Prof. Jian Yang. His research interests encompass computer vision and machine learning, with a focus on 3D-related tasks including depth estimation, depth completion, and reconstruction. He has published over ten papers in top conferences such as CVPR, ICCV, and NeurIPS.
\end{IEEEbiography}
\vskip -0.32in
\begin{IEEEbiography}
    [{\includegraphics[width=1in,height=1.25in,clip,keepaspectratio]{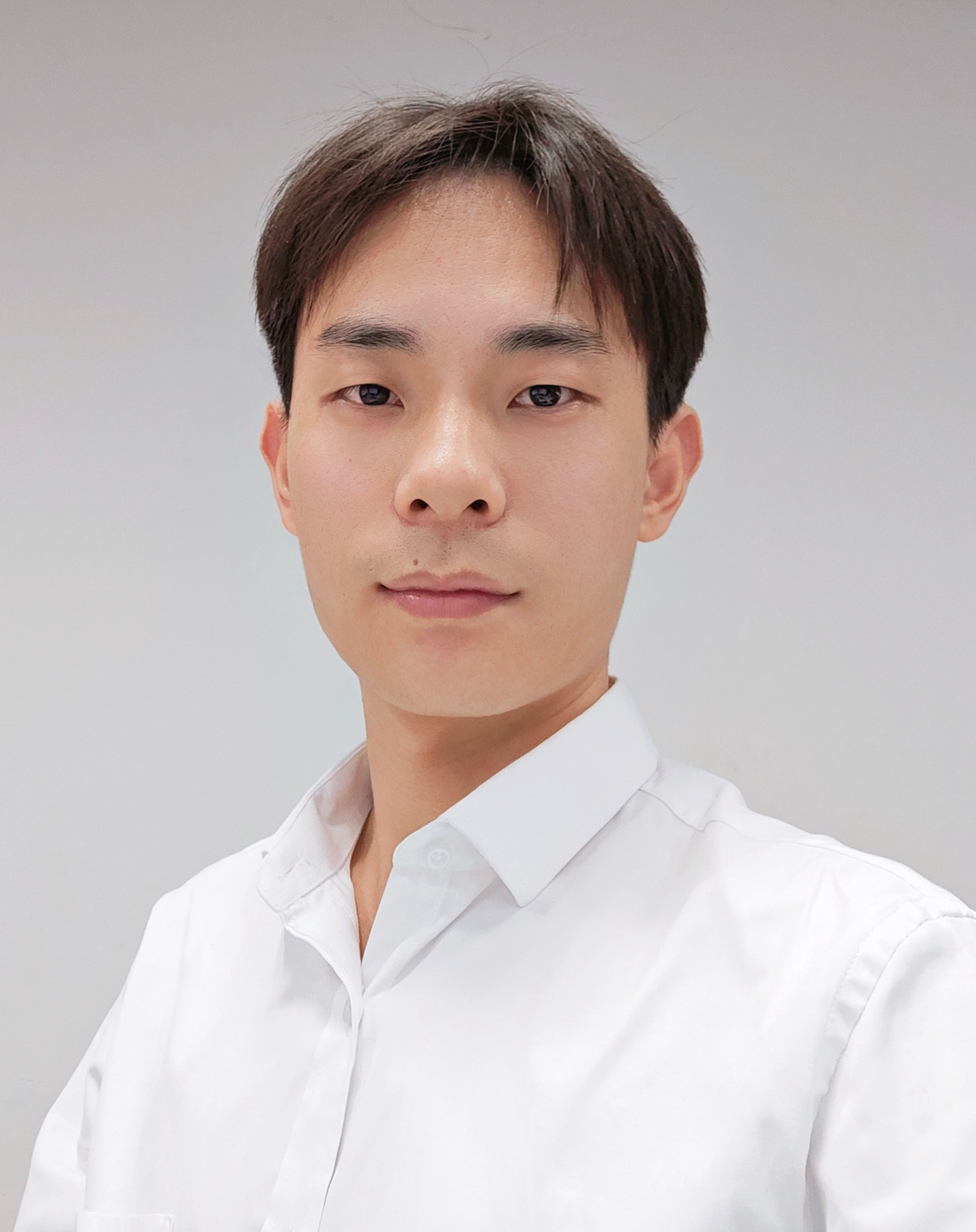}}]{Zhiqiang Yan} received the Ph.D. degree from the Nanjing University of Science and Technology, Nanjing, Jiangsu, China, in 2024, supervised by Prof. Jian Yang. His research interests include computer vision and machine learning, especially on the tasks of depth estimation, depth completion, and depth super-resolution, all of which are crucial for 3D reconstruction, autonomous driving, and other related 3D visual perception. He has published 10+ papers in top conferences such as CVPR and ICML. 
\end{IEEEbiography}
\vskip -0.32in
\begin{IEEEbiography}[{\includegraphics[width=1in,height=1.25in,clip,keepaspectratio]{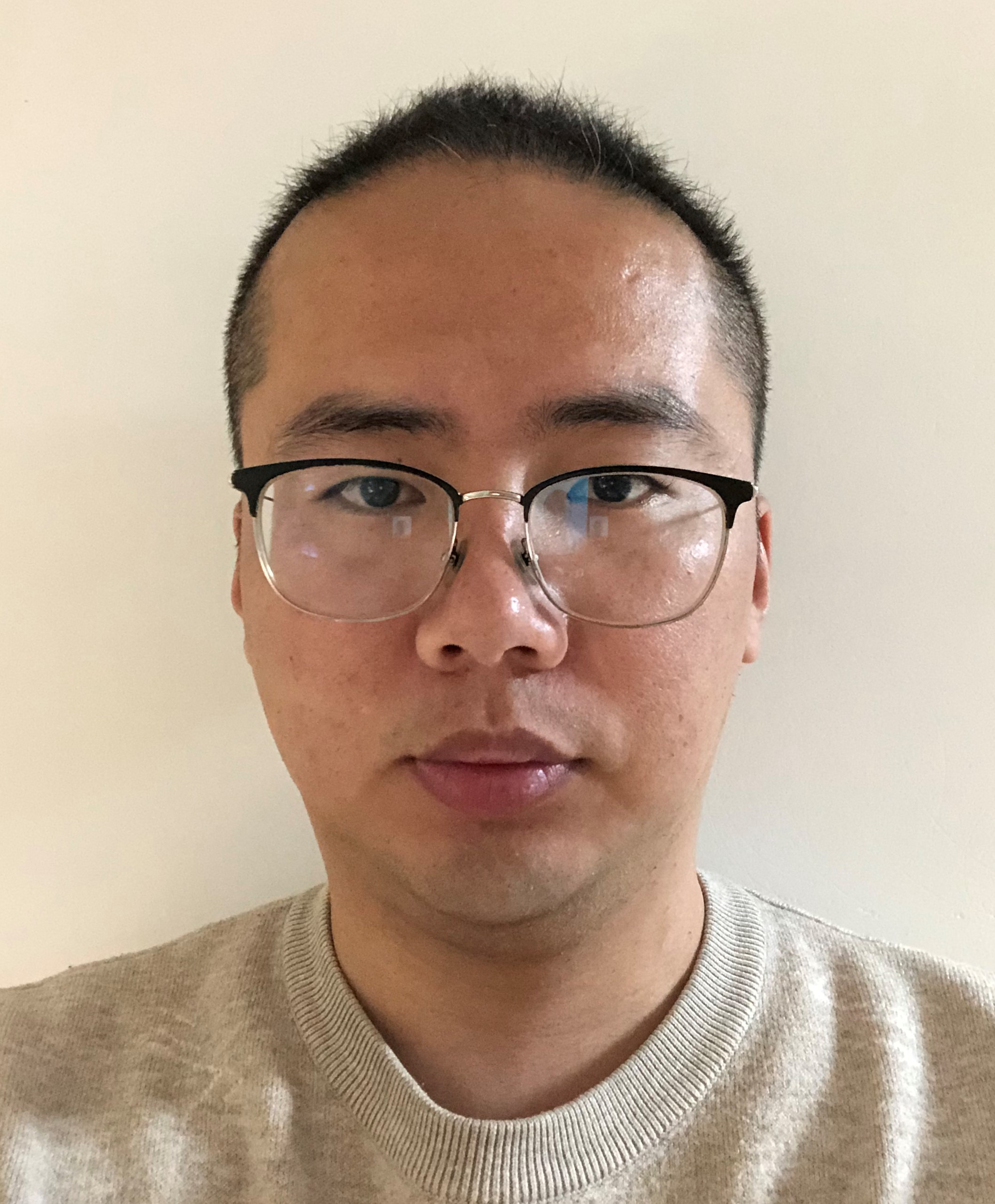}}]{Junkai Fan} received his B.E. degree in the Department of Computer Science and Technology at Fuyang Normal University, Fuyang, Anhui, China, in 2015. He is currently pursuing a Ph.D. degree in Computer Science and Engineering at Nanjing University of Science and Technology, Nanjing, China. His research interests include computer vision, with a focus on real-world image restoration and depth estimation. He has published numerous papers in top conferences such as CVPR, AAAI, NeurIPS.
\end{IEEEbiography}
\vskip -0.32in
\begin{IEEEbiography}[{\includegraphics[width=1in,height=1.25in,clip,keepaspectratio]{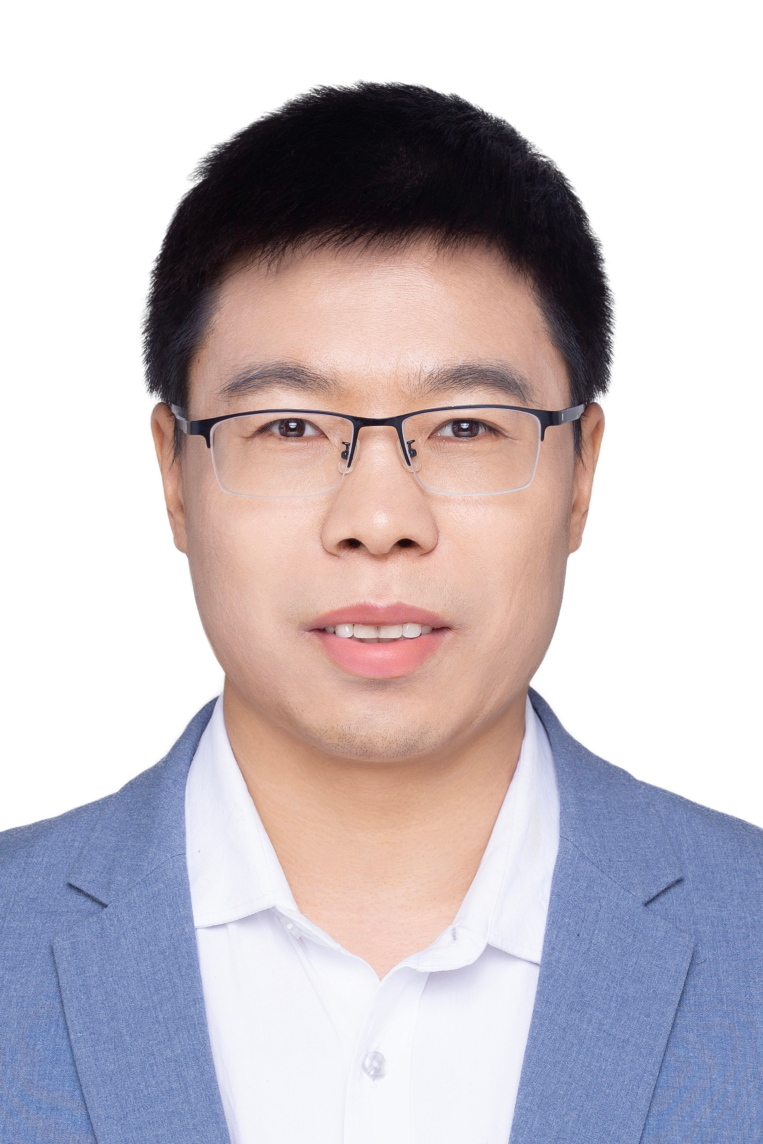}}]{Jun Li (M'16)} received the Ph.D. degree in pattern recognition and intelligence systems from the Nanjing University of Science and Technology, Nanjing, China, in 2015. From October 2012 to July 2013, he was a Visiting Student at the Department of Statistics, Rutgers University, Piscataway, NJ, USA. From December 2015 to October 2018, he was a Post-Doctoral Associate with the Department of Electrical and Computer Engineering, Northeastern University, Boston, MA, USA. From November 2018 to October 2019, he was a Post-Doctoral Associate with the Institute of Medical Engineering and Science, Massachusetts Institute of Technology, Cambridge, MA, USA. He has been a Professor with the School of Computer Science and Engineering, Nanjing University of Science and Technology, since 2019. His research interests are computer vision and creative learning. Dr. Li has served as an SPC/PC Member for CVPR/ICCV/ECCV/NeurIPS/ICML/ICLR/AAAI, and a reviewer for many international journals, such as the IEEE TRANS. ON Intell. Transp. Syst./TRANS. ON IMAGE PROCESSING/TRANS. ON CYBERNETICS.
\end{IEEEbiography}

\begin{IEEEbiography}[{\includegraphics[width=1in,height=1.25in,clip,keepaspectratio]{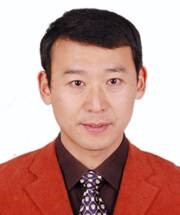}}]{Jian Yang} received the Ph.D. degree from Nanjing University of Science and Technology (NUST) in 2002, majoring in pattern recognition and intelligence systems. From 2003 to 2007, he was a Postdoctoral Fellow at the University of Zaragoza, Hong Kong Polytechnic University and New Jersey Institute of Technology, respectively. From 2007 to present, he is a professor in the School of Computer Science and Technology of NUST. Currently, he is also a visiting distinguished professor in the College of Computer Science of Nankai University. His papers have been cited over 50000 times in the Scholar Google. His research interests include pattern recognition and computer vision. Currently, he is/was an associate editor of Pattern Recognition, Pattern Recognition Letters, IEEE Trans. Neural Networks and Learning Systems, and Neurocomputing. He is a Fellow of IAPR.  
\end{IEEEbiography}

%
%
%
%
%

\vfill

\end{document}